\documentclass{article}
\usepackage{amssymb}

\usepackage[final]{corl_2025} %

\usepackage[numbers]{natbib}
\usepackage{multicol}
\usepackage{multirow}

\usepackage{hyperref}
\hypersetup{
	colorlinks=true,
	linkcolor=orange,
	filecolor=magenta,      
	urlcolor=orange,
	citecolor=orange,
}
\usepackage{url}
\usepackage{float}

\usepackage{mathtools}
\usepackage{makecell}
\usepackage{amsfonts}  %
\usepackage{graphics}
\usepackage[pdftex]{graphicx}
\usepackage{wrapfig}
\usepackage{caption}
\usepackage{color}
\usepackage{dsfont}
\usepackage[]{mdframed}
\usepackage{algorithm}
\usepackage[noend]{algpseudocode}
\usepackage{xparse}
\usepackage{amsmath}
\usepackage{bm}
\usepackage{subcaption}
\usepackage{mathtools}
\usepackage{amssymb}
\usepackage{framed}
\usepackage{amsthm}
\usepackage{amssymb}
\usepackage{cleveref}

\usepackage{booktabs}
\usepackage{epigraph}

\usepackage{xcolor} %
\usepackage{hyperref} %
\usepackage{booktabs}
\usepackage{multirow}
\usepackage{graphicx}
\usepackage{tabularx}
\usepackage{listings}

\hypersetup{
    colorlinks=true,
}

\usepackage{colortbl} %
\definecolor{lightblue}{RGB}{217, 237, 247} %
\definecolor{lightgreen}{RGB}{229, 245, 224}

\usepackage{xargs} %
\usepackage[colorinlistoftodos,prependcaption,textsize=tiny]{todonotes}
\newcommandx{\wrn}[2][1=]{\todo[linecolor=red,backgroundcolor=red!25,bordercolor=red,#1]{#2}}
\newcommandx{\cmt}[2][1=]{\todo[linecolor=blue,backgroundcolor=blue!25,bordercolor=blue,#1]{#2}}

\algrenewcommand{\algorithmiccomment}[1]{\hfill\textcolor{lightgray}{$\triangleright$\;#1}}

\renewcommand{\cref}{\Cref}  %

\usepackage{listings}
\usepackage{xcolor}
\lstdefinelanguage{Python}{
  morekeywords=[1]{True, False, None, and, as, assert, async, await, break, class, continue, def, del, elif, else, except, finally, for, from, global, if, import, in, is, lambda, nonlocal, not, or, pass, raise, return, try, while, with, yield},
  morekeywords=[2]{transformers, numpy},
  sensitive=true,
  morecomment=[l]{\#},
  morestring=[b]',
  morestring=[b]",
}

\definecolor{darkgreen}{RGB}{0,128,0}
\definecolor{darkblue}{RGB}{0,0,255}
\definecolor{darkred}{RGB}{139,0,0}

\def \Acronym {RoboArena}
\def \NUniversities {7}
\def \NPolicies {7}
\def \NEvals {4284}
\def \NPairwise {612}

\newcommand{\website}{\url{https://robo-arena.github.io}}

\title{\Acronym: Distributed Real-World Evaluation of Generalist Robot Policies}

\author{
  \hspace{-2cm}Pranav Atreya$^{\ast, 1}$ \quad Karl Pertsch$^{\ast, 1, 2}$ \quad Tony Lee$^{\ast, 2}$ \\
  \hspace{-2cm}\textbf{Moo Jin Kim}$^{2}$ \quad \textbf{Arhan Jain}$^{3}$ \quad \textbf{Artur Kuramshin}$^{4}$ \quad \textbf{Clemens Eppner}$^{5}$ \quad \textbf{Cyrus Neary}$^{4}$ \quad \textbf{Edward Hu}$^{6}$\\
  \hspace{-2cm}\textbf{Fabio Ramos}$^{5}$ \quad \textbf{Jonathan Tremblay}$^{5}$ \quad \textbf{Kanav Arora}$^{3}$ \quad \textbf{Kirsty Ellis}$^{4}$ \quad \textbf{Luca Macesanu}$^{7}$ \quad \textbf{Marcel Torne Villasevil}$^{2}$ \\
  \hspace{-2cm} \textbf{Matthew Leonard}$^{6}$ \quad \textbf{Meedeum Cho}$^{8}$ \quad \textbf{Ozgur Aslan}$^{4}$ \quad \textbf{Shivin Dass}$^{7}$ \quad \textbf{Jie Wang}$^{6}$ \quad \textbf{William Reger}$^{7}$ \\
  \hspace{-2cm}\textbf{Xingfang Yuan}$^{6}$ \quad \textbf{Xuning Yang}$^{5}$ \quad \textbf{Abhishek Gupta}$^{3}$ \quad \textbf{Dinesh Jayaraman}$^{6}$ \\
  \hspace{-2cm}\textbf{Glen Berseth}$^{4}$ \quad \textbf{Kostas Daniilidis}$^{6}$ \quad \textbf{Roberto Martin-Martin}$^{7}$ \quad \textbf{Youngwoon Lee}$^{8}$\\
  \hspace{-2cm}\textbf{Percy Liang}$^{2}$ \quad \textbf{Chelsea Finn}$^{2}$ \quad \textbf{Sergey Levine}$^{1}$\\[0.3cm]
  \hspace{-2cm}\website
}

\begin{document}
\maketitle

\renewcommand*{\thefootnote}{\fnsymbol{footnote}}
\footnotetext{
{\fontsize{8pt}{11pt}\selectfont
$^\ast$: Core contributors, detailed contributions in \cref{sec:contributions}\\Correspondence to: \href{mailto:pranavatreya@berkeley.edu,pertsch@berkeley.edu,tonyhlee@stanford.edu}{\texttt{pranavatreya@berkeley.edu, pertsch@berkeley.edu, tonyhlee@stanford.edu}}\\
    $^1$UC Berkeley, $^2$Stanford University, $^3$University of Washington, $^4$University of Montreal, $^5$NVIDIA, $^6$University of Pennsylvania, $^7$UT Austin, $^8$Yonsei University
}
}
\renewcommand*{\thefootnote}{\arabic{footnote}}

\begin{abstract}

Comprehensive, unbiased, and comparable evaluation of modern generalist policies is uniquely challenging: existing approaches for robot benchmarking typically rely on heavy standardization, either by specifying fixed evaluation tasks and environments, or by hosting centralized ``robot challenges'', and do not readily scale to evaluating generalist policies across a broad range of tasks and environments. In this work, we propose RoboArena, a new approach for scalable evaluation of generalist robot policies in the real world. Instead of standardizing evaluations around fixed tasks, environments, or locations, we propose to crowd-source evaluations across a distributed network of evaluators. Importantly, evaluators can freely choose the tasks and environments they evaluate on, enabling easy scaling of diversity, but they are required to perform double-blind evaluations over \emph{pairs} of policies. Then, by aggregating preference feedback from pairwise comparisons across diverse tasks and environments, we can derive a ranking of policies. We instantiate our approach across a network of evaluators at seven academic institutions using the DROID robot platform. Through more than 600~pairwise real-robot evaluation episodes across seven generalist policies, we demonstrate that our crowd-sourced approach can more accurately rank the performance of existing generalist policies than conventional, centralized evaluation approaches, while being more scalable, resilient, and trustworthy. We open our evaluation network to the community and hope that it can enable more accessible comparisons of generalist robot policies.

\end{abstract}


\begin{figure}[t]
    \centering
    \includegraphics[width=\linewidth]{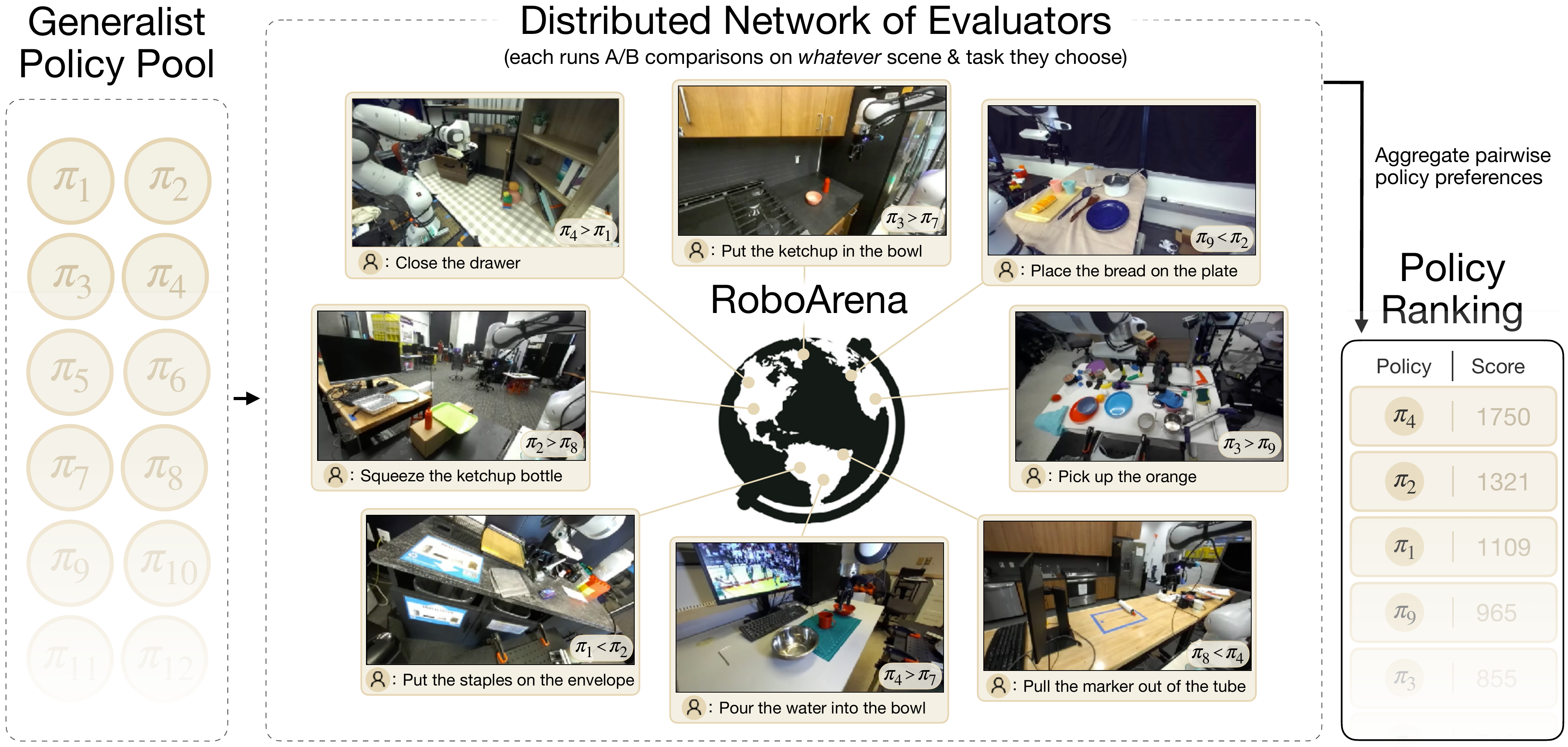}
    \caption{We present \Acronym{}, a distributed real-world evaluation framework for generalist robot policies. Instead of standardizing environments and tasks, \Acronym{} aggregates crowd-sourced pairwise A/B policy evaluations across a broad spectrum of environments and tasks to derive a global policy ranking. Its decentralized design makes \Acronym{} a scalable, comprehensive, and trustworthy framework for generalist robot policy evaluation. We open-source an instantiation of \Acronym{} on the DROID robot platform~\citep{khazatsky2024droid} and invite community members to participate, both by contributing policies and running evaluations.
    }
    \label{fig:teaser}
\end{figure}

\section{Introduction}
\label{sec:intro}

Modern robot policies are increasingly general, and able to perform a wide range of tasks across many environments~\citep{open_x_embodiment_rt_x_2023, kim2024openvla, etukuru2024robotutilitymodelsgeneral, pertsch2025fast, geminirobotics2025}. With the increased generality a new challenge arises: how can we accurately measure and compare the performance of such generalist policies? Conventional approaches to robot evaluation and benchmarking rely on comparing policies across tightly standardized sets of environments and tasks~\citep{calli2015ycb, yang2019replab, heo2023furniturebench, luo2023fmb}, and are typically restricted to only a handful of tasks across a small number of scenes~\citep{krotkov2018darpa, bauer2022real, zhou2023train, yenamandra2024towards}. As such, they are not well-suited to provide comprehensive performance evaluations for policies that are designed to perform across a much broader spectrum of initial conditions. At the same time, existing evaluation approaches are challenging to scale: guaranteeing comparable conditions across a large number of real-world tasks and environments faces many practical challenges, from accurately reproducing scene layouts and lighting conditions, to maintaining large, centralized fleets of evaluation robots, and manufacturing differences between said robots. As such, the development of increasingly general robot policies requires us to rethink the way we evaluate robot policies.

At first glance, broadly capable robot policies seem to compound the reproducibility and scalability challenges faced by robot evaluations and benchmarks in the past. 
Yet, in this work, we argue that increasingly general robot policies offer an \emph{opportunity} for an alternative approach to robot evaluation, that has the promise to address many of the challenges of prior evaluation efforts. %
Realizing this potential, however, requires rethinking how we run robot evaluations: instead of \emph{standardization} and centralized evaluation ``challenges'', we argue that \emph{decentralization} and an evaluation approach that embraces the non-stationarity of the physical world offer a promising alternative. %

To this end, we propose \Acronym{}, a distributed real-world evaluation framework for generalist robot policies. Inspired by recent crowdsourced benchmarks for evaluating generalist language or vision-language models like Chatbot Arena~\citep{chiang2024chatbot} or GenAI Arena~\citep{jiang2024genai}, \Acronym{} relies on a decentralized network of evaluators that perform \emph{pairwise}, \emph{double-blind} comparisons of policies in \emph{whichever scene} and on \emph{whatever task} they deem suitable. The evaluator then provides a preference for which of the two policies performed better, along with a free-form language explanation. Our evaluation algorithm aggregates a large number of such pairwise comparisons into a global policy ranking, as well as a set of qualitative characteristics, strengths and weaknesses for each policy. This decentralized design allows \Acronym{} to be \textbf{open-ended} in the number of environments and tasks policies are evaluated on, \textbf{robust}, since many entities across the decentralized network contribute via double-blind evaluations to the final score, and \textbf{scalable}, with dozens of evaluators asynchronously contributing policy evaluations at any time of day. By not relying on identical initial conditions beyond the horizon of a single pairwise policy comparison, \Acronym{} foregoes many of the practical challenges prior robot evaluation frameworks faced.

In this paper, we outline our distributed evaluation protocol, including algorithms for aggregating pairwise policy comparisons into global policy rankings, and for extracting qualitative policy characteristics via LLM-assisted analysis of evaluation results. We then instantiate \Acronym{} on the DROID robot platform~\citep{khazatsky2024droid}, on which modern policies can out-of-the-box generalize to new scenes and tasks~\citep{pertsch2025fast}. Through decentralized evaluations of \NPolicies{}~generalist policies across \NUniversities{}~universities and a total of \NPairwise{}~pairwise comparisons, we demonstrate that \Acronym{} provides more accurate performance rankings of generalist policies than the conventional, centralized evaluation scheme used in prior work, when compared to an ``oracle'' policy ranking computed via exhaustive evaluation of all policies on all tested tasks. At the same time, \Acronym{} matches the episode efficiency of standard evaluation, i.e., the same number of evaluations, when distributed across the \Acronym{} evaluator network, lead to higher quality policy rankings. In addition to the evaluation framework, our work also provides the most comprehensive evaluations of generalist robot policies to date (\NEvals{}~evaluation episodes across numerous tasks and scenes), and highlights the limitations of current policies.

\section{Related Work}
\label{sec:related}

\paragraph{Simulated and real-world robot evaluation.}
Simulated evaluations offers perfect reproducibility, and can be efficiently parallelized. Thus, numerous simulated robot benchmarks have been proposed over the years~\citep{tassa2018deepmind, todorov2012mujoco, kolve2017ai2, james2020rlbench, lee2021ikea, mees2022calvin, srivastava2022behavior, liu2023libero, bao2023dexart, xiang2023rmbench, pumacay2024colosseum, nasiriany2024robocasa}. However, simulated environments are an imperfect replica of the real world, and even works that optimize for visual and physical fidelity~\citep{li24simpler} are often limited in the diversity of tasks they support, and sometimes fail to accurately reflect real-world policy performance~\citep{zhou2025autoeval}. In the real world, benchmarks try to ensure reproducibility through detailed manuals for reproducing environment setups~\citep{calli2015ycb, yang2019replab, heo2023furniturebench, luo2023fmb, collins2023ramp, khargonkar2024scenereplica}, enabling remote access to centrally hosted evaluations~\citep{pickem2017robotarium, bauer2022real, zhou2023train, zhou2025autoeval}, or organizing in-person challenges~\citep{kitano1997robocup, krotkov2018darpa, correll2016analysis, yenamandra2024towards, EarthRoverChallengeWebsite}. However, to ensure reproducibility of initial conditions and make centralized evaluation feasible, these approaches typically restrict evaluations to a narrow set of tasks and environments, or even a specific challenge date. %
In this work, we propose an alternative approach for real-world robot evaluation based on crowd-sourcing pairwise policy comparisons across a \emph{decentralized} network of evaluators. Our approach can cover a wider distribution of tasks and environments, and is inherently more resilient and scalable than existing, centralized evaluation approaches.

\paragraph{Generalist robot policies.} 
Recently, robotics has seen a trend towards training \emph{generalist} robot policies~\citep{driess2023palm, rt22023arxiv, open_x_embodiment_rt_x_2023, octo_2023, Doshi24-crossformer, kim2024openvla, black2024pi_0, wen2024tinyvlafastdataefficientvisionlanguageaction, zhen20243dvla, belkhale2024minivla, szot2024multimodal, pertsch2025fast, geminirobotics2025, wen2025dexvla, bjorck2025gr00t} on diverse datasets of robot experience~\citep{dasari2019robonet, ebert2021bridge, walke2023bridgedata, open_x_embodiment_rt_x_2023, khazatsky2024droid, bharadhwaj2023roboagent, fang2024rh20t, shafiullah2023bringingrobotshome, contributors2025agibotworld}. Notably, multiple works have demonstrated policies that can perform tasks across many environments \emph{out of the box}~\citep{gupta2018robotlearninghomesimproving, chi2024universal, etukuru2024robotutilitymodelsgeneral, pertsch2025fast}. At the same time, conventional approaches to robot evaluation struggle to comprehensively evaluate the performance of such generalist policies, since they are cumbersome to scale to broad distributions of tasks and environments, motivating our distributed evaluation approach in this work.

\paragraph{Crowd-sourced benchmarks.} 
In recent years, crowd-sourced benchmarks have gained popularity, from language modeling~\citep{chiang2024chatbot}, to generative image and video modeling~\citep{jiang2024genai}, or even for specialized legal model training~\citep{guha2023legalbench}. In robotics, \citet{dasari2022rb2} proposed a benchmark that aggregates policy ranking results across institutions to obtain more comprehensive evaluations of learning \emph{algorithms}, but it required retraining of policies for every new institution and task at hand, and thus remained limited to a small number of tested environments and tasks. In contrast, our work proposes a distributed benchmark for \emph{generalist robot policies}, which can be evaluated out-of-the-box, and thus makes it feasible to evaluate on a much broader distribution of tasks and environments. %

\section{Decentralized Robot Policy Evaluation via Pairwise Comparison}
\label{sec:method} 

Conventional robot policy evaluations aim to \emph{standardize} the conditions under which policies are compared as much as possible, typically by defining fixed sets of tasks and scenes policies should be run on, and by trying to closely match initial conditions like lighting, camera angles, or scene background~\citep{calli2015ycb, heo2023furniturebench, walke2023bridgedata, kress2024robot}. As we discussed in \cref{sec:intro}, reproducing such standardized conditions in the real world is extremely challenging, particularly when evaluations should be run not just in one central place, but across institutions. 

In this section, we introduce an alternative approach for robot evaluation that foregoes repeatedly evaluating policies in standardized settings, and instead relies on \emph{decentralized, pair-wise, double-blind policy comparisons}. Crucially, our evaluation approach \emph{does not} prescribe what scene or task a policy pair should be evaluated in, but instead aggregates a large number of pairwise comparisons across \emph{many} tasks and scenes to establish a policy ranking. This decentralized approach has multiple benefits:
it is \textbf{open-ended}, since we do not standardize tasks and environments, thus broadening coverage; it is \textbf{robust}, since no single entity can easily sway results in double-blind, decentralized evaluations; it is \textbf{scalable}, since many institutions can collaborate on the evaluations; and it is \textbf{adaptable}, since tasks can naturally adapt to the frontier of policy capabilities.

In the remainder of this section, we will first describe our evaluation procedure~(\cref{sec:eval_procedure}), then detail how we aggregate pairwise comparisons into global policy rankings~(\cref{sec:policy_ranking}), %
and finally describe tools for extracting qualitative policy characteristics from the evaluation data~(\cref{sec:qualitative_analysis_tools}).

\subsection{Policy Evaluation Protocol}
\label{sec:eval_procedure}

\begin{wrapfigure}{r}{0.6\textwidth}
\begin{minipage}{0.6\textwidth}
\vspace{-0.8cm}
\begin{algorithm}[H]
\caption{Policy Evaluation Protocol}
\label{alg:eval_protocol}
\begin{algorithmic}[1]
\Require Set of policies $\Pi$, Evaluator $E$, Central Server $C$
\Ensure Global policy ranking, policy characteristics

\For{$i = 1$ to $K$ evaluations}
    \State $E$ requests policies for evaluation from $C$
    \State $C$ samples two policies $\pi_A, \pi_B \sim \Pi$ %
    \State $E$ rearranges scene and defines task $T_i$
    \State $E$ executes $\pi_A$ and $\pi_B$ sequentially
    \State $E$ provides pairwise feedback $F_i(\pi_A, \pi_B)$ to $C$
\EndFor
\State $C$ aggregates $\{F_i\}_{i=1}^K$ into global ranking \Comment{Section 3.2}
\State $C$ extracts qualitative policy characteristics \Comment{Sec. 3.3}
\end{algorithmic}
\end{algorithm}
\end{minipage}
\vspace{-0.5cm}
\end{wrapfigure}
We provide a summary of our policy evaluation protocol in \cref{alg:eval_protocol}. Given a pool of policies $\pi_{1 \dots N} \in \Pi$, our goal is to estimate a performance ranking. %
We design our evaluation for \emph{generalist} robot policies, and thus assume that all policies in $\Pi$ can be meaningfully (and safely) evaluated across a broad range of scenes and tasks. Additionally, we assume access to a pool of evaluators $E$ that asynchronously run real robot evaluations, and a central evaluation server $C$ that manages the decentralized evaluation operation.

During an evaluation session, an evaluator $E$ requests two policies from the central server $C$. $C$ randomly samples two policies $(\pi_A, \pi_B) \in \Pi$ and assigns them to $E$. 
To ensure unbiased evaluation, the evaluators do not know which policies they are evaluating. In practice, we simply provide them with the IP addresses of remotely hosted evaluation servers. After policies are sampled, the evaluator arranges the evaluation scene, e.g., by moving the robot to a new location and rearranging the objects in front of the robot, and defines the evaluation task $T_i$ in form of a natural language instruction. Then, $E$ runs rollouts for policies $\pi_A$ and $\pi_B$ back-to-back until the task is completed or a fixed timeout is reached. Importantly, we require the evaluator $E$ to \emph{closely match} the initial conditions \emph{within} this A/B policy comparison (while they can choose to change them \emph{between} separate pairwise evaluations). This ensures that the comparison of policies $\pi_A$ and $\pi_B$ is fair.

After both evaluations are complete, $E$ provides feedback $F(\pi_A, \pi_B)$ about the performance of the policies. \textbf{We ask evaluators to provide three types of feedback:} a continuous \textbf{progress score} $\in [0 \dots 100]$ that is proportional to the maximum \emph{progress} a policy achieved on the task (e.g., 0 for no progress, 100 for successfully executing the task, intermediate values for partial success); a binary, \textbf{pairwise preference} label that indicates which policy the evaluator preferred (we leave it to the evaluator to decide how to determine their policy preference); and a free-form, \textbf{natural language explanation} for \emph{why} they preferred one policy over the other. 

After the task instruction $T$, pairwise feedback $F$, and recordings of all observations and actions are sent to the central server $C$, the evaluator may choose to continue with another evaluation session, or pause and return at a later time. All evaluations outside a single pairwise comparison can run fully \emph{asynchronously} at any time or place. %

\subsection{Computing Global Policy Rankings}
\label{sec:policy_ranking}
In this section, we discuss our algorithm for computing a global policy ranking using the pairwise feedback provided by the evaluators. Formally, we are given a set of $N$ policies $\Pi = \{\pi_1, \dots, \pi_N\}$ and a dataset of pairwise preferences $\mathcal{D}_p = \{P_{\pi_A, \pi_B}, t\}$, where $P_{\pi_A, \pi_B} \in \{0, 1\}$ indicates a binary preference, and $t$ identifies the task the A/B evaluation was run on (e.g., a specific scene and language instruction). We aim to compute a global policy ranking $\mathcal{R}: \pi_i > \pi_j > \dots > \pi_k$. 

\textbf{Bradley-Terry Model~~~} The Bradley-Terry (BT) model~\citep{bradley1952rank} is commonly used in a learning from preferences setting. BT models the win-probability of policy $\pi_A$ versus $\pi_B$ as the sigmoid of the difference of log-abilities: $p(\pi_A > \pi_B) = \sigma(\theta_A - \theta_B)$. Then, the log-ability parameters of this model can be fit either via an online algorithm such as Elo~\citep{elo1967proposed}%
, or in an offline setting with an iterative algorithm guaranteed to converge to the maximum likelihood fit of the data~\cite{zermelo1929berechnung}. This iterative algorithm can also be modified to support ties~\cite{davidson1970extending}. 

\textbf{Extending BT~~~} The Bradley-Terry model, while appealing due to the existence of stable offline solvers, assumes that each pairwise comparison in the preference dataset takes place under identical conditions. This assumption is not satisfied in the setting of A/B robot policy evaluations, where the \emph{task} under which the A/B evaluation is conducted can vary from one evaluation to the next. Importantly, the task can significantly affect the policy preference: for example, a task that is very difficult or very easy for both policies can diminish any differentiating signal between $\pi_A$ and $\pi_B$, and thus the probability that $\pi_A$ is preferred over $\pi_B$ by evaluators. 
Further, pairs of policies can exhibit different relative performance relationships on different subsets of tasks; a specialized policy for one subset of tasks would, for example, be preferred over more generalist policies on this subset, but not in aggregate. Thus, it is important to account for task-effects when modeling the preference data. Standard BT models would simply treat task-related effects as noise, leading to worse rankings.

We propose to augment the BT log-ability parameters for each of the policies $\theta = (\theta_1 \dots \theta_N)$ with task difficulty parameters $\tau = (\tau_1 \dots\ \tau_T)$, marginal task probabilities $\nu = (\nu_1 \dots \nu_T)$ s.t. $\sum_t \nu_t = 1$, defining the prior probability any given A/B trial belongs to latent bucket $t$,
and policy-task offsets $\psi = ((\psi_{1_1} \dots \psi_{1_T}), \dots, (\psi_{N_1} \dots \psi_{N_T}))$, which model \emph{policy-dependent} task difficulty:
\vspace{-1em}
\begin{equation}
    p(\pi_A > \pi_B) = \sum_{t=1}^{T} \nu_t \cdot \sigma(\theta_A + \psi_{A_t} - \tau_t) \cdot (1 - \sigma(\theta_B + \psi_{B_t} - \tau_t))
\end{equation}
Critically, all parameters can be learned solely from preference data; no auxiliary task information is required. Via the maximum likelihood estimation process, the task-related parameters $\tau, \nu$, and $\psi$ will be automatically fit. The number of task buckets $T$ is a hyperparameter.%

\textbf{An EM algorithm for approximate MLE~~~} The parameters $\theta, \tau, \nu$ and $\psi$ can be fit via an approximate maximum likelihood expectation-maximization (EM) algorithm. In essence, the algorithm iterates between measuring the likelihood of the data under the current model parameters, calculating the first and second-order derivatives of this likelihood, performing a maximization step with clipped Newton updates, and centering the new parameters to maintain zero mean. A detailed derivation of the derivatives and description of the EM update algorithm can be found in \cref{app:ranking_model}.

\subsection{Extracting Qualitative Policy Characteristics}
\vspace{-2mm}
\label{sec:qualitative_analysis_tools}

\begin{figure}
    \centering
    \includegraphics[width=\linewidth]{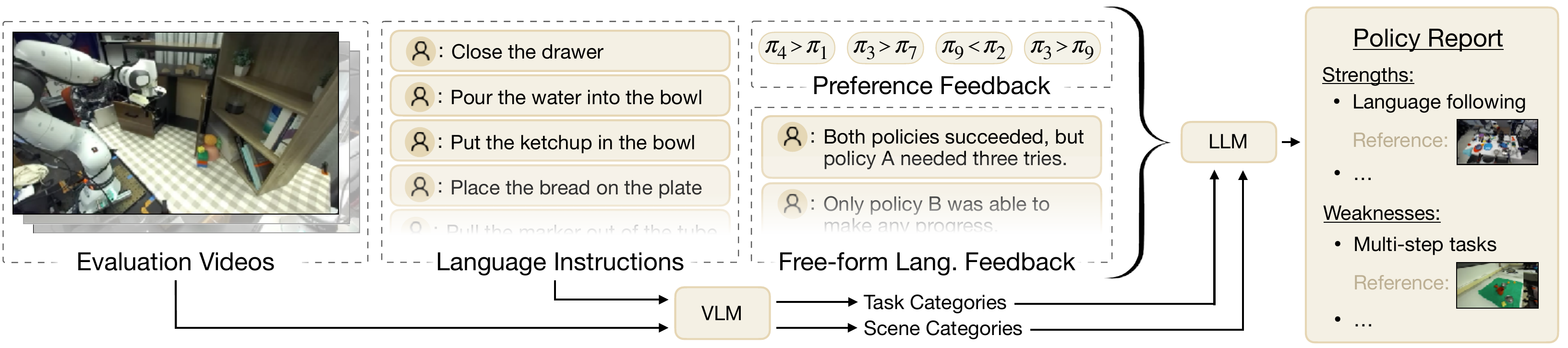}
    \caption{Pipeline for extracting \emph{qualitative} policy characteristics from \Acronym{}'s rich evaluation data. We use a VLMs to categorize scenes and tasks, and then use an LLM to aggregate information across a large number of evaluation rollouts into a \emph{policy report} that summarizes qualitative strengths and weaknesses, and cites concrete evaluation rollout videos as evidence.
    }
    \label{fig:analysis_pipeline}
\end{figure}

Estimating \emph{qualitative} policy characteristics, e.g., a model's ability to follow language or perform multi-step tasks, is crucial to guide future research. Typically, researchers develop an intuition for such qualitative characteristics by running all evaluations for a given policy themselves. Yet, in a distributed evaluation setup we need new tools to synthesize such nuanced insights from hundreds of policy rollout videos, task instructions, and evaluator language feedback. To this end, we experiment with using large language and vision-language models (LLMs, VLMs) to assist with the analysis.

We provide an overview of our prototype analysis tool in \cref{fig:analysis_pipeline}. We pass the first images of each evaluation video and the corresponding task instruction to a VLM  (\texttt{OpenAI GPT‑4.5}) and ask it to categorize the task (e.g., pick-place vs. open-close vs. tool use) and describe the scene's lighting, clutter, object visibility, etc. Then, for each policy in our pool, we use an LLM  (\texttt{OpenAI o3}) to generate a \emph{policy report} by summarizing preference annotations, categorization results, and free-form evaluator feedback for all evaluations. We instruct the LLM to compare performance to other policies along the task categories, and to extract qualitative policy characteristics from the language feedback. Importantly, the LLM is instructed to \emph{cite} evaluation episodes as evidence for any claim in the report and we automatically annotate the report with videos from these rollouts to enable researchers to verify any claims. We experimentally verify our tool in \cref{sec:qualitative_exp}, and include example policy reports in the supplementary data.

\section{The DROID-\Acronym{} Evaluation System}
\label{app:droid_eval_system}

\begin{wrapfigure}{r}{0.4\textwidth}
    \centering
    \vspace{-0.5cm}
    \includegraphics[width=\linewidth]{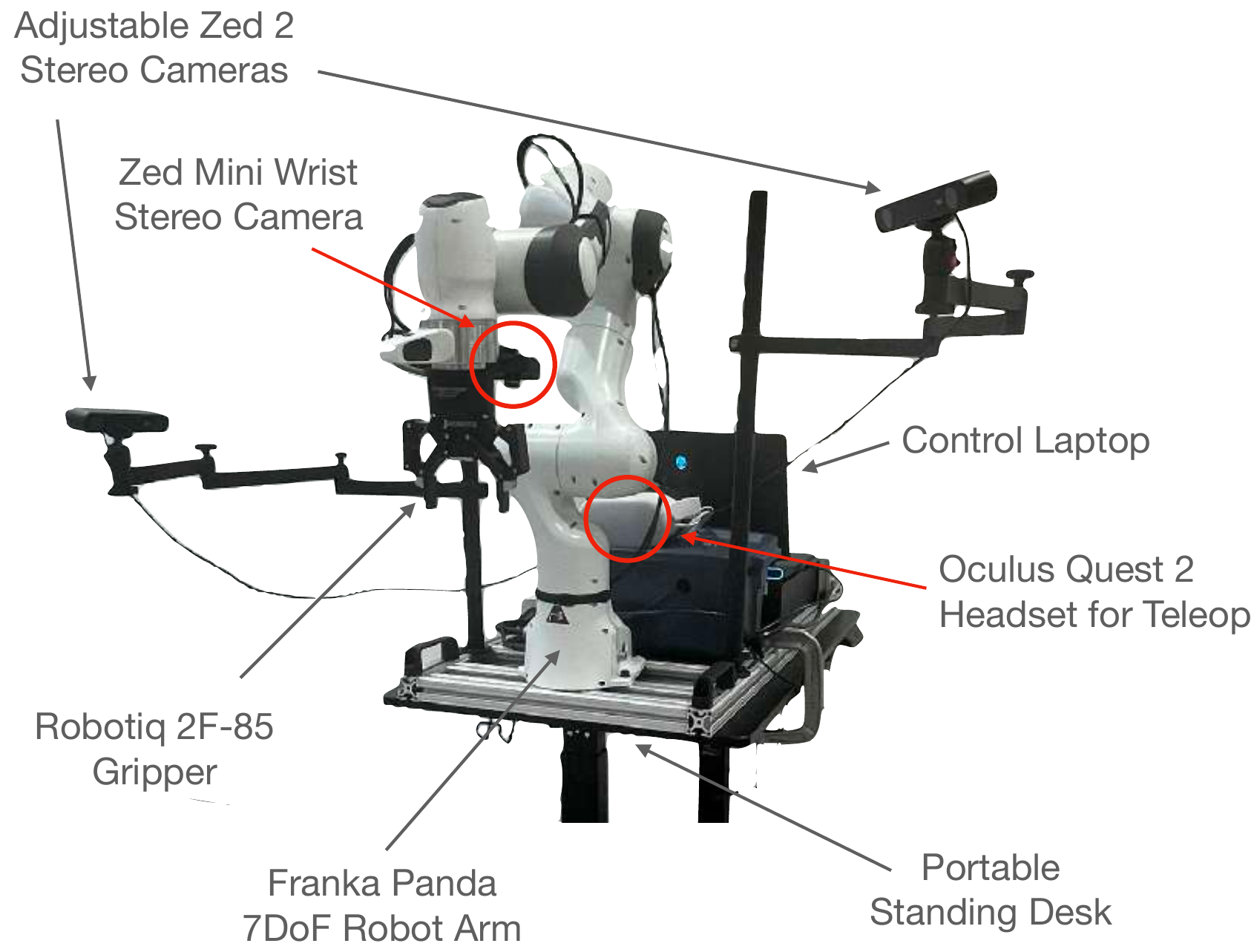}
    \caption{The DROID robot setup, which we use for the DROID-\Acronym{} evaluation system. Reproduced with permission from \citet{khazatsky2024droid}.
    }
    \label{fig:droid_platform}
    \vspace{-1cm}
\end{wrapfigure}

We instantiate \Acronym{} in a prototype evaluation system on the DROID robot platform~\citep{khazatsky2024droid}, which consists of a Franka Panda 7DoF robot arm, a Robotiq 2F-85 parallel-jaw gripper, a ZED-mini stereo wrist camera and one or multiple external ZED~2 stereo cameras (see~\cref{fig:droid_platform}). We choose the DROID platform since it offers multiple attractive properties:
\begin{itemize}
    \item The Panda arm provides sufficient dexterity to perform a wide range of manipulation tasks across varied real-world environments.  
    \item The robot is mounted on a height-adjustable, mobile table, which enables rapid reconfiguration of scenes and camera viewpoints.  
    \item Most importantly, the platform is associated with the open-source DROID dataset~\citep{khazatsky2024droid}, a large-scale, real-world dataset that supports training of \emph{generalizable} robot policies.
    \item Finally, DROID setups are already deployed across multiple academic institutions worldwide, which is key to enabling distributed evaluation at scale.
\end{itemize}

We note that, while DROID is a convenient platform to develop a prototype distributed robot evaluation, the \Acronym{} evaluation approach readily extends to other robot embodiments and potentially even to evaluating cross-embodiment policies~\citep{open_x_embodiment_rt_x_2023} across platforms. The remainder of this section, we provide details on our evaluation system design (\cref{app:system_design}) and outline safety considerations and incentive mechanisms that support open participation from the broader robotics community (\cref{sec:open_sourcing}).

\subsection{\Acronym{} System Design}
\label{app:system_design}

\begin{figure}
    \centering
    \includegraphics[width=\linewidth]{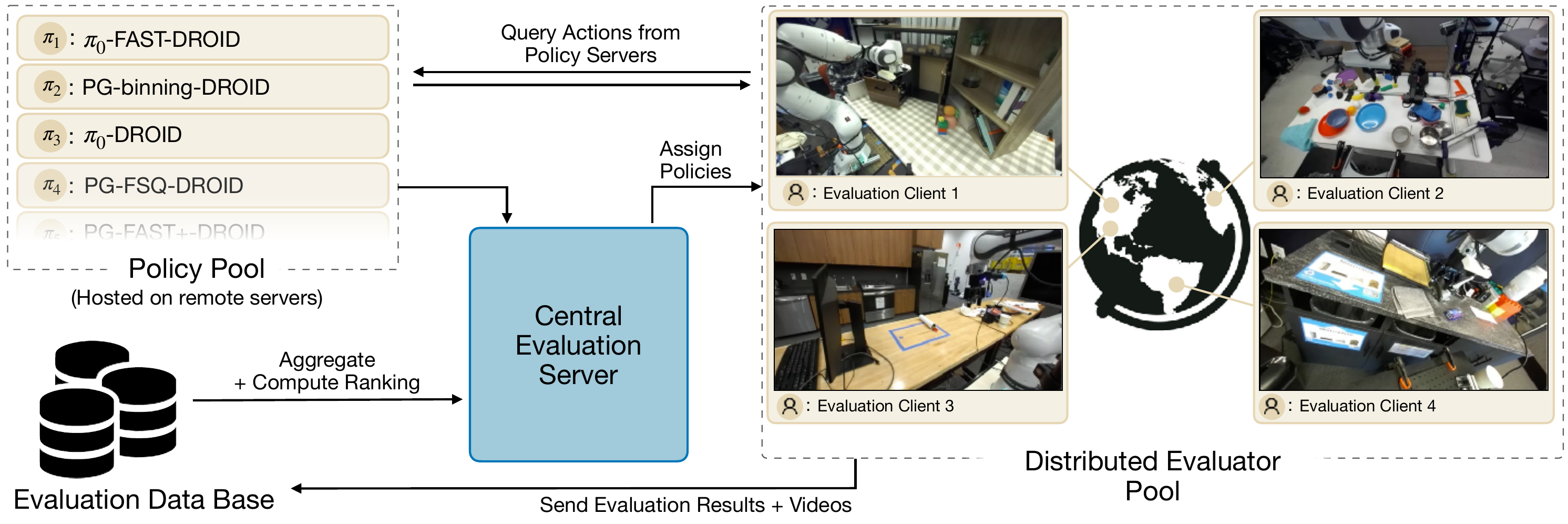}
    \caption{The DROID-\Acronym{} system consists of a pool of remotely hosted policy servers, a pool of distributed evaluator ``clients'' with real robot setups, a database for storing evaluation results, and a central evaluation management server that orchestrates communication, aggregates the evaluation results, and computes a policy ranking.
    }
    \label{fig:system_overview}
\end{figure}

We design a system for performing decentralized evaluation over a potentially large pool of policies, with a pool of evaluators that may span many different locations, countries, or even continents. Our system has four core components: policy inference servers, evaluation clients, an evaluation database, and a central evaluation server (see \cref{fig:system_overview}).

\textbf{Policy inference servers~~~} We host all policies in our pool on remote servers, instead of running them ``on-premise'' at the robot evaluation station. This has multiple benefits: \emph{effective resource utilization} since multiple evaluators can share the same policy server, a \emph{lightweight evaluation client} since no inference compute needs to be provided client-side, which lowers the barrier for contributing evaluations, and \emph{user-side control} over policy servers since we assume users host their own policies when submitting them for evaluation, which ensures that policies are run correctly, equipped with sufficient inference compute, and proprietary models can be evaluated without sharing model weights. Remote hosting may, however, introduce additional latency during policy inference, but in practice we found this to be negligible for the static manipulation tasks on which we typically evaluate DROID policies.
We acknowledge that future versions of \Acronym{} evaluations may need to support (at least partially) local inference to enable evaluation of more dynamic tasks.

\textbf{Evaluation clients~~~} We provide a client-side script that guides users through the evaluation protocol outlined in \cref{sec:eval_procedure}. It handles communication with the central evaluation server and the policy inference servers. Since no client-side inference compute is required, this script can be run on any computer that is connected to the Internet and the DROID robot.

\textbf{Evaluation database~~~} This database hosts all evaluation results: instructions, scores and preference, natural language feedback, and the rollout videos. Information is uploaded by the clients.

\textbf{Central evaluation server~~~} We host a central evaluation server that assigns policies to evaluators, and keeps track of newly registered or deprecated policies in the policy pool and automatically cancels evaluations after a timeout, e.g., if evaluation clients crashed or lost network connection. %

\subsection{Open-Sourcing \Acronym{}: Interfaces, Safety and Incentives}
\label{sec:open_sourcing}

Our goal in designing the DROID-\Acronym{} is to provide it as a resource to the robotics community, to which anyone can submit policies and contribute evaluations. Importantly, DROID-\Acronym{} can enable researchers without access to a physical robot to train and evaluate real-world generalist robot policies -- now all required resources are open-source: diverse, real-robot datasets~\citep{walke2023bridgedata, open_x_embodiment_rt_x_2023, khazatsky2024droid, contributors2025agibotworld}, scalable modeling frameworks~\citep{octo_2023, kim2024openvla, ren2023openpizero, pi2025openpi, bjorck2025gr00t}, and DROID-\Acronym{} real-robot evaluations.

Making the DROID-\Acronym{} easy to use, safe, and self-sustaining requires a few additional elements. First, we make it easy to browse all performed evaluations and view existing policy leaderboards on our website. We plan to publish separate leaderboards for ``all policies'' and ``open-source policies''; the latter are trained only on publicly available datasets, and provide weight access and details for how to reproduce training. Additionally, we design multiple \textbf{safety layers} to prevent policies from damaging robot evaluation hardware: first, we test that any newly submitted policy server complies with the expected input and output formats. Then we run said policy in a ``test environment'' across a few different scenes and tasks with a specifically trained evaluator that is able to quickly intervene if a policy runs the danger of acting unsafely and damaging the robot (akin to a test driver for autonomous vehicles). Only after a policy passed these tests, we add it to the general pool and distribute it to evaluators.

Finally, to make the DROID-\Acronym{} self-sufficient in the long term, we implement an \textbf{``evaluation credit'' system}, that balances evaluation supply and demand: for every pairwise policy evaluation that an evaluator runs, they receive a credit, which they can use to request an equal number of pairwise comparisons between their own policies and other policies from the pool. This way, evaluation effort for any individual is comparable to running evaluations for just their own policy on their own setup, but by agreeing to evaluate others' policies through the decentralized benchmarking system, they effectively get a much broader evaluation coverage for the same effort. To support participation of researchers without access to a physical robot (and thus without the ability to contribute evaluations themselves), multiple institutions agreed to ``sponsor'' a base budget of evaluations that will be distributed among all users ``free of charge''. 

\section{Experiments}
\label{sec:experiments}

The goal of our experimental evaluation is to answer the following questions: (1)~How does \Acronym{} compare to conventional robot evaluation approaches for ranking the performance of generalist robot policies? (2)~How \emph{sample efficient} is \Acronym{} in ranking robot policies? (3)~Can \Acronym{} extract \emph{qualitative} insights about policy performance beyond aggregate performance?

\subsection{Experimental Setup}

\textbf{Policy pool.} We populate the policy pool with all publicly available \emph{generalist} DROID policies that have been shown to work out of the box in new environments. At the time of writing and to our knowledge, there are \emph{seven} such policies, based off PaliGemma~\citep{beyer2024paligemma} or $\pi_0$~\citep{black2024pi_0} base models. %
Concretely, we evaluate the following policies: \textbf{$\pi_0$-flow-DROID}, the $\pi_0$ flow-vision-language-action model (VLA)~\citep{black2024pi_0}, fine-tuned on the DROID dataset; \textbf{$\pi_0$-FAST-DROID}, the $\pi_0$-FAST autoregressive VLA~\citep{pertsch2025fast}, fine-tuned on the DROID dataset; \textbf{PG-flow-DROID, PG-FAST-DROID, PG-FAST+-DROID, PG-FSQ-DROID, PG-Bin-DROID}, PaliGemma~\citep{beyer2024paligemma} vision-language models (VLMs), fine-tuned on the DROID dataset using different action representations from \citet{pertsch2025fast}: $\pi_0$-style flow matching, FAST/FAST+ tokenization~\citep{pertsch2025fast}, finite scalar quantization~\citep{fsq}, and simple binning tokenization~\citep{rt22023arxiv, kim2024openvla}. We use the publicly available implementation in \citep{pi2025openpi} for all policies.

\textbf{Comparison.} We establish an ``oracle'' policy ranking by exhaustively evaluating \emph{all} policies on \emph{all} tested tasks and comparing their average progress scores. This was done by asking evaluators to score the remaining five policies after each A/B evaluation on the same task. In total, we use \NEvals{}~evaluations to compute this oracle ranking. We compare \Acronym{} to a conventional robot policy evaluation approach, which tests all policies on a fixed, tightly standardized set of evaluation tasks in a narrower set of environments. \textbf{Concretely, we compare to the DROID evaluation procedure used in \citet{pertsch2025fast}}, which consists of 17~tasks and 44~episodes per policy and is representative of typical robot evaluations (\citep{kim2024openvla, Zawalski24-ecot}, see the appendix of \citet{pertsch2025fast} for a detailed list of tasks). %

\subsection{\Acronym{} Accurately Ranks Policy Performances}
\label{sec:ranking_results}
\vspace{-1em}

\begin{figure}
    \centering
    \includegraphics[width=\linewidth]{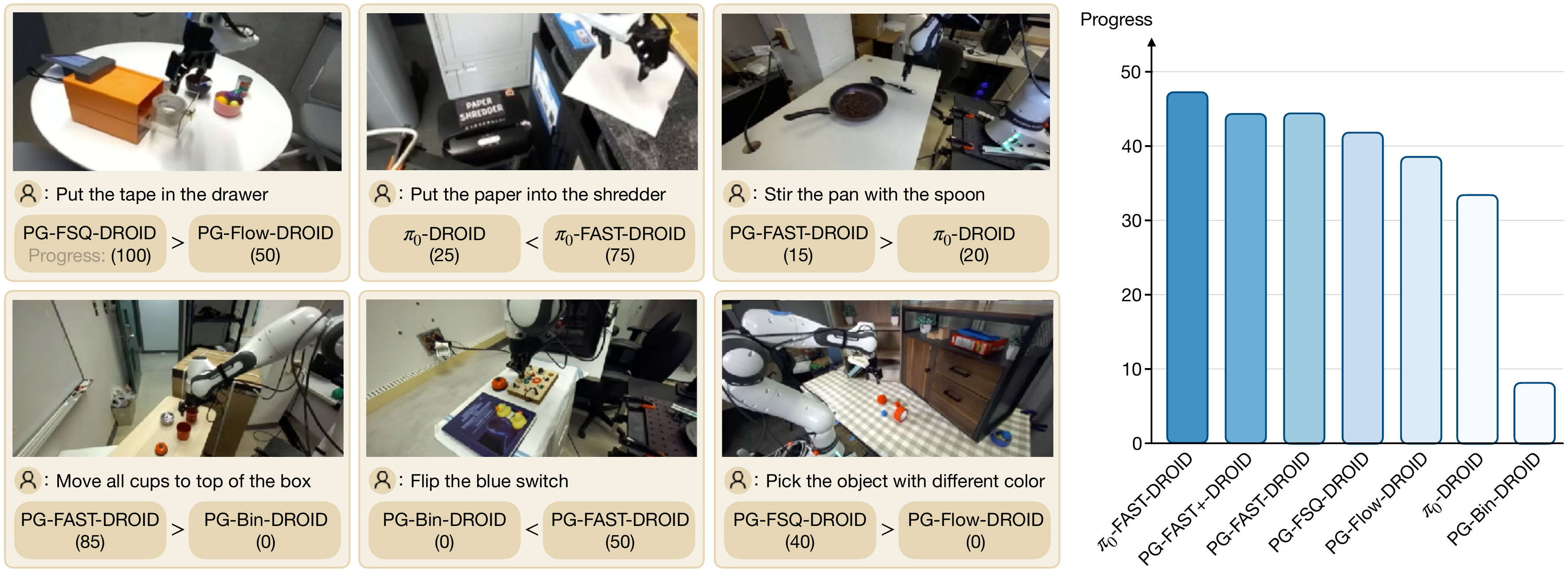}
    \caption{\textbf{Left}: Examples of \Acronym{} evaluations. Evaluations span a diverse set of scenes and tasks. \textbf{Right}: ``Oracle'' policy ranking, aggregated from progress scores of \NEvals{}~evaluation rollouts. 
    }
    \label{fig:ranking_result}
\end{figure}

\begin{wrapfigure}{R}{0.5\textwidth}
    \centering
    \vspace{-0.3cm}
    \includegraphics[width=0.5\textwidth]{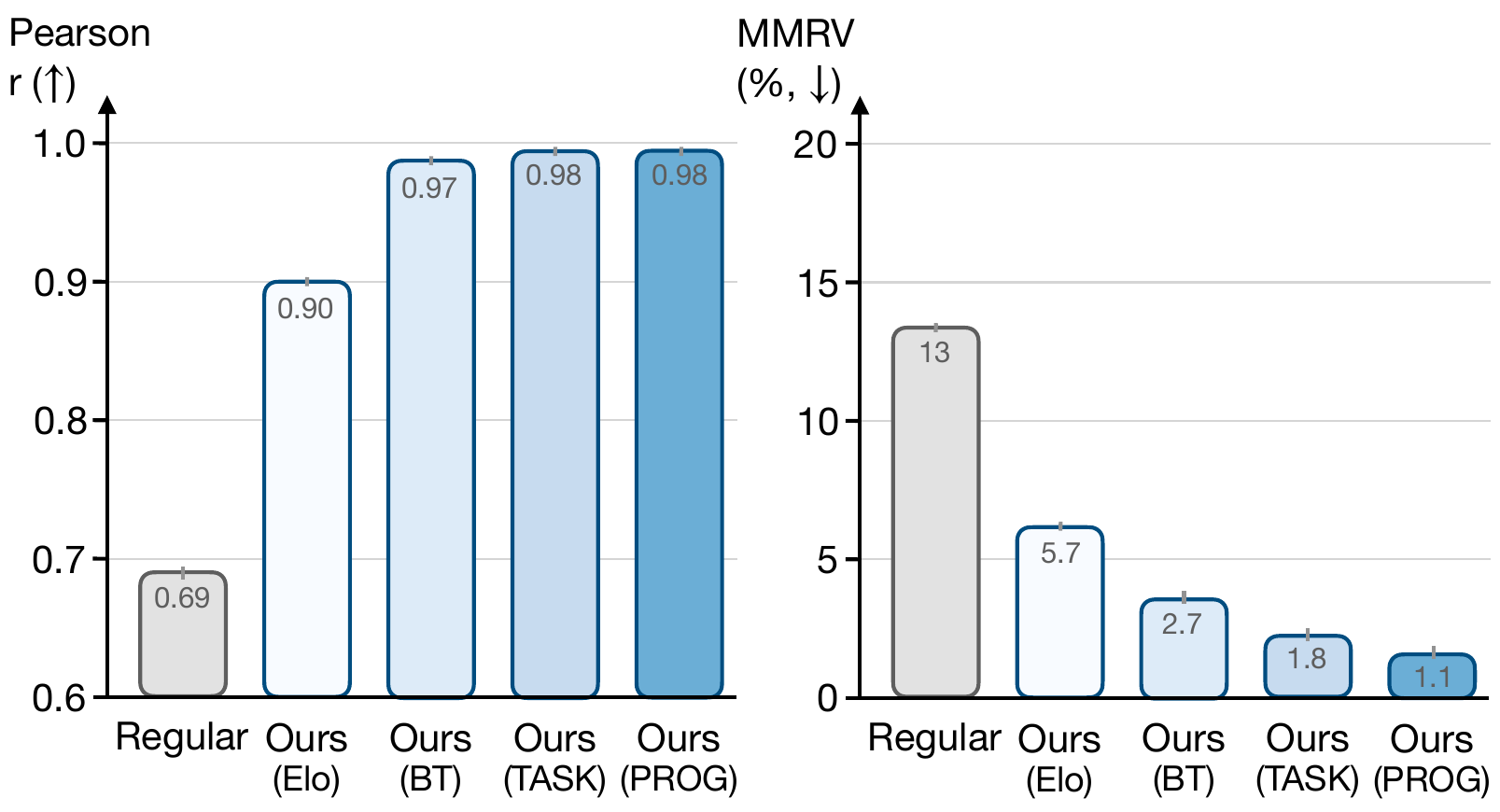}
    \caption{
        Policy rankings from \Acronym{} pairwise comparisons correlate significantly better with oracle rankings than conventional robot evaluation approaches (``Regular''). Our task-aware ranking approach (``TASK'') leads to the best ranking. 
        Ranking with progress scores (``PROG'') also proves effective, but may lose more nuanced information about policy performance. %
    }
    \vspace{-0.5cm}
    \label{fig:correlation_results}
\end{wrapfigure}
In total, our \Acronym{} evaluation covers \NEvals{}~policy rollouts across dozens of scenes and hundreds of task instructions. We show a few examples in \cref{fig:ranking_result}, left, and many more in \cref{app:detailed_eval_breakdown}. To our knowledge, this is the most extensive evaluation of generalist policies to date. While the remainder of this section is devoted to evaluating our \Acronym{} evaluation framework itself, we provide a detailed analysis of the underlying evaluation data, including strengths and weaknesses of current generalist policies, in \cref{app:policy_eval_summary}.

We evaluate how well our pairwise evaluation approach can approximate the oracle ranking (see \cref{fig:ranking_result}, right), which is computed through exhaustive evaluation of all policies on all tasks. We compare \Acronym{} to conventional robot evaluations from prior work. We follow \citet{li24simpler} and report Pearson correlation $r$, as well as Mean Maximum Rank Violation (MMRV), a ranking metric that takes the \emph{performance difference} between policies into account.

We test instantiations of \Acronym{} with standard ranking algorithms, namely Elo~\citep{elo1967proposed} and Bradley-Terry (``BT'', \citep{bradley1952rank}), and our task-aware ranking approach introduced in \cref{sec:policy_ranking} (``TASK''). The results in \cref{fig:correlation_results} show that conventional robot evaluations (``Regular''), which due to reproducibility challenges are restricted to a relatively small number of tasks and environments, do not provide a reliable performance estimate for \emph{generalist} policies. \textbf{This highlights the benefit of evaluations \emph{without} task and environment standardization, as they lead to much greater diversity, and thus effective ranking for generalist policies (\cref{fig:correlation_results}).} Additionally, we find that our task-aware ranking approach leads to more accurate rankings than both standard Elo computation and the conventional Bradley-Terry model. 
The resulting rankings of both, oracle and our approach follow the intuitions of prior work: expressive action representations outperform simple binning tokenization~\citep{pertsch2025fast, lee2024behavior, black2024pi_0, chi2023diffusion}, and discrete action tokenization (``FAST'', ``FSQ'') outperform diffusion policies (``Flow'', ``$\pi_0$-DROID'') in language-conditioned evaluations~\citep{pertsch2025fast, intelligence2025pi} (\cref{fig:ranking_result}). 

We also test using an average of the progress scores for each of the pairwise policy comparisons to compute a global ranking (``PROG''). The results in \cref{fig:correlation_results} show that this strategy achieves good correlation to the oracle as well. However, ranking policies based on progress scores alone can miss more nuanced feedback on policy performance. We find that evaluators regularly score both policies in an A/B comparison with the \emph{same} progress score, yet express clear \emph{preference} for one policy over the other, e.g., because it acted more swiftly or confidently (see \cref{app:comp_over_progress} for examples). Thus, progress-based and preference-based rankings are complementary, and should both be reported.%

\subsection{\Acronym{} Evaluation is Sample Efficient}

\begin{wrapfigure}{R}{0.5\textwidth}
\vspace{-0.5cm}
    \centering
    \includegraphics[width=0.5\textwidth]{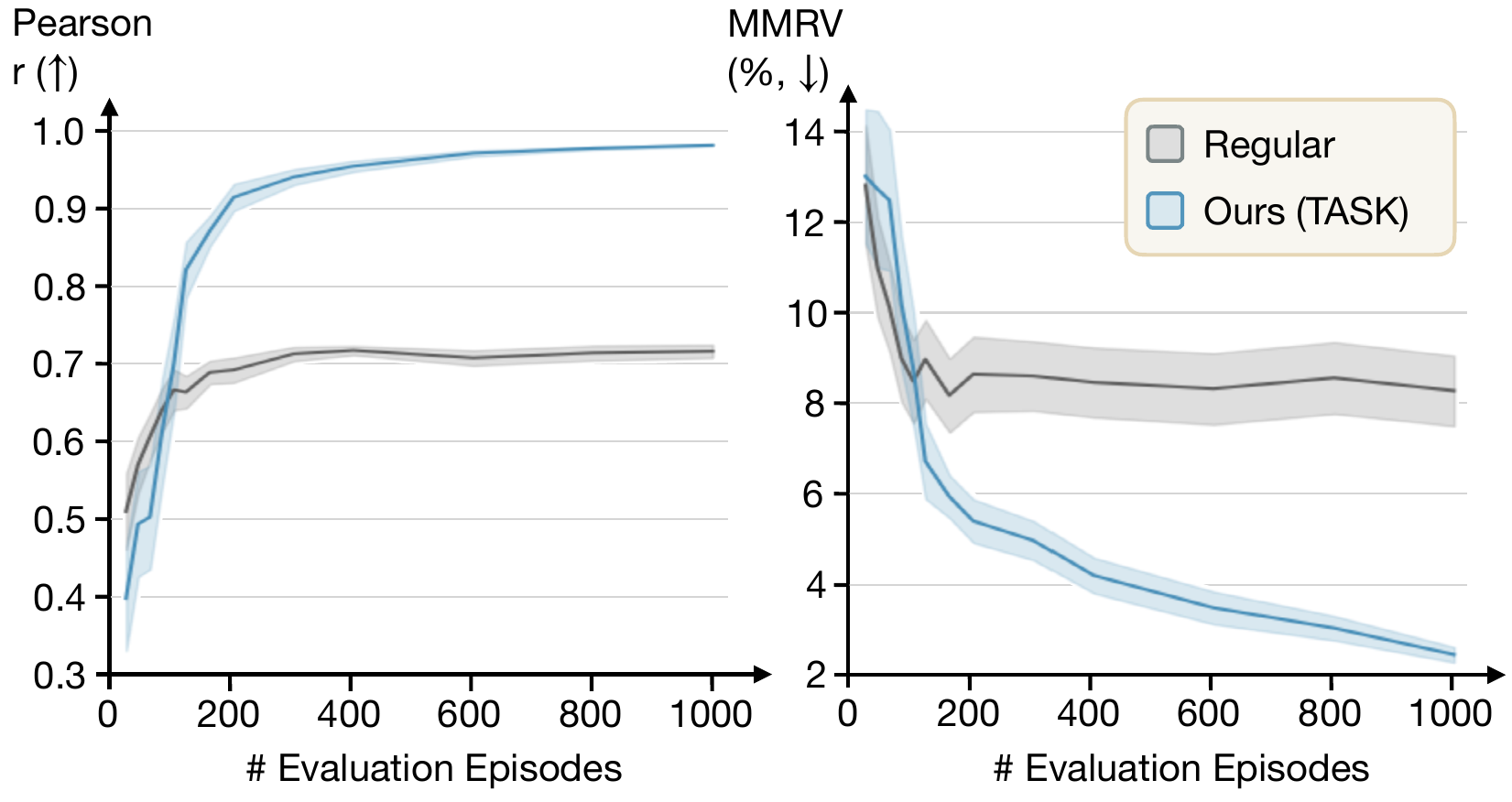}
    \caption{
        Rank correlation as a function of number of evaluation episodes. \Acronym{} converges to high-quality rankings within just 100~pairwise comparisons. %
    }
    \vspace{-0.3cm}
    \label{fig:convergence_results}
\end{wrapfigure}
We investigate how many evaluations are required to produce an accurate ranking using \Acronym{}: we compute rankings for differently-sized, random subsets of our full evaluation data and report ranking accuracy as a function of the number of evaluation episodes in \cref{fig:convergence_results}. We find that \Acronym{} converges to high-quality rankings within just 100~pairwise comparisons, matching the convergence speed of conventional robot evaluations, while providing significantly more accurate rankings. The quality of \Acronym{} rankings further improves as more comparisons are collected. This suggests, that distributed \Acronym{} evaluations offer an appealing alternative to regular policy evaluations.

\subsection{Extracting qualitative policy characteristics}
\label{sec:qualitative_exp}

\begin{wrapfigure}{r}{0.35\textwidth}
\vspace{-1cm}
\begin{tabularx}{0.35\textwidth}{l|X}
\toprule
Model & Task Category Acc. \\
\midrule
GPT-4o & 94.6\% \\
\bottomrule
\end{tabularx}
\caption{Task categorization accuracy (448~samples).}
\label{tab:categorization_performance}
\vspace{-0.4cm}
\end{wrapfigure}
In this section, we evaluate the quality of the LLM/VLM-assisted \emph{policy reports} (\cref{sec:qualitative_analysis_tools}). %
First, we evaluate the VLM category predictions by comparing them to category assignments made by a human expert, and we find that they are approximately 95\% accurate (\cref{tab:categorization_performance}).

\begin{wrapfigure}{R}{0.5\textwidth}
    \centering
    \vspace{-0.5cm}
    \includegraphics[width=0.5\textwidth]{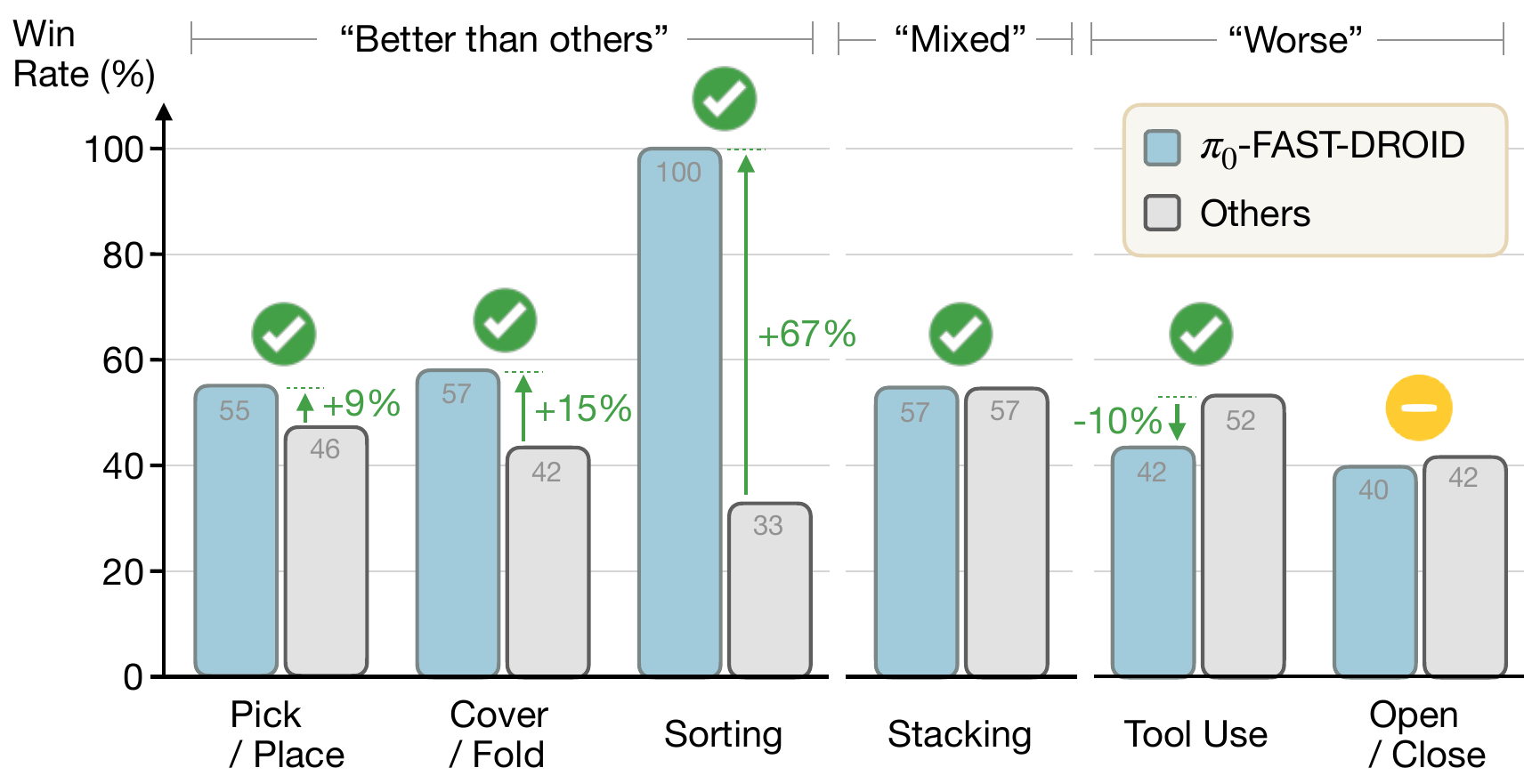}
    \caption{
        Claims made by our LLM-assisted analysis tools about \texttt{$\pi_0$-FAST-DROID} outperforming, matching, or underperforming other policies are supported by the win rates in the evaluation data. 
    }
    \vspace{-0.3cm}
    \label{fig:cat_results}
\end{wrapfigure}

Next, we examine whether the comparative claims in the generated policy report align with the evaluation data, using the example of the strongest policy in our pool: $\pi_0$-FAST-DROID~\citep{pertsch2025fast}. Concretely, the report compares $\pi_0$-FAST-DROID's performance to that of other policies in the pool along different task categories. In \cref{fig:cat_results} we show, that for most categories for which the report claims that $\pi_0$-FAST-DROID outperforms, matches, or underperforms other policies, these claims are supported by the respective win rates in the evaluation data. %
We encourage readers to review the full, interactive report with video references in our supplementary material. %

\section{Discussion}
\label{sec:conclusion} 
\vspace{-1em}
We introduced \Acronym{}, a distributed framework for evaluating generalist robot policies. We have shown that by aggregating evaluations across a decentralized network of evaluators, each running pairwise policy comparisons on many different tasks and scenes, \Acronym{} can generate more accurate policy performance rankings than conventional, centralized evaluation approaches, while retaining high evaluation sample efficiency. We have also introduced prototype tools for LLM-assisted analysis of the evaluation results. We will open-source our \Acronym{} evaluation framework and give other researchers access for contributing policies and evaluations, in an effort to make evaluations of generalist robot policies more comparable.

\section{Limitations}
\label{sec:limitations}

\textbf{Cross-embodiment.} While \Acronym{} is a general approach for robot evaluation, the experimental evaluations in this paper have focused on the DROID platform~\citep{khazatsky2024droid}, since it was well-suited for developing our evaluation framework. However, there is a growing interest in developing \emph{cross-embodiment} policies, that can not only operate across many scenes and tasks, but also across robot embodiments. Future work should investigate how \Acronym{} can be extended to diverse robot embodiments and still support policies that may only be evaluatable on specific embodiments.

\textbf{Controlled experimentation.} The design of \Acronym{}, which is focused on decentralized evaluation without restrictions on tasks or scenes, makes it challenging to perform experiments that only vary \emph{a single} condition at a time (e.g. only camera angle, or only object position). As such, \Acronym{} is complementary to targeted, smaller-scale evaluations that focus on individual axes of generalization~\citep{gao2025taxonomy}. %

\textbf{Adversarial evaluators.} While \Acronym{}'s distributed, double-blind evaluation scheme gives it an inherent robustness against individual influencing, we have not investigated its robustness to intentionally adversarial evaluators that try to temper with evaluation results, for example by providing random preference ratings or intentionally misleading language feedback. Future work should investigate how distributed robot evaluation approaches can be hardened against such tampering.

\textbf{Over-optimization and Gotthart's Law.} A common wisdom is that a measure, e.g., for model performance, ceases to be a good measure when it becomes a target. This potential for over-optimization, also known as Gotthart's law, is innate to any measure, and has recently been shown to also impact crowd-sourced evaluations~\citep{theregister2025llamacheating}, even if they may inherently be more robust to such over-optimization than static benchmarks. While in robotics, the current (limited) performance of policies makes such over-optimization less likely, future work should critically examine whether evaluations with approaches like \Acronym{} remain well-correlated with perceived real-world policy performance.

\acknowledgments{We thank Siyi Huang, Ellie Huynh, Emiliano Adrian Sanchez, Justin Kang, and Sarah Kunda for help with evaluating policies. We also thank Kyle Stachowicz for help with repairing the DROID robot station during development. This research was partly supported by RAI, ONR N00014-25-1-2060, ONR N00014-22-1-2621, ONR N00014-22-12677NSF, NFS IIS-2150826, NSF CAREER 2239301, NSF 2331783, DARPA HR00112490428, DARPA TIAMAT HR00112490421, the Amazon Science Hub, the Toyota Research Institute, the Army Research Lab, and the National Research Foundation of Korea (NRF) grant funded by the Korean government (MSIT) (RS-2024-00333634). We acknowlegde compute support from the Google TPU Research cloud (TRC). We also want to acknowledge funding support from Natural Sciences and Engineering Research Council of Canada, Fonds de recherche du Québec and The Canadian Institute for Advanced Research (CIFAR) and compute support from Digital Research Alliance of Canada, Mila IDT, and NVidia. We would also like to thank John Edwards Leadership Fund (CFI) for funding the purchase of part of the hardware for this project.}

\bibliography{references}  %

\clearpage
\appendix

\section{Contributions}
\label{sec:contributions}

\textbf{\Acronym{} system design and implementation}: Pranav Atreya, Karl Pertsch, Tony Lee

\textbf{Experiment design and analysis}: Pranav Atreya, Karl Pertsch, Tony Lee

\textbf{Policy evaluation}: Pranav Atreya, Tony Lee, Moo Jin Kim, Karl Pertsch, Arhan Jain, Artur Kuramshin, Cyrus Neary, Edward Hu, Kanav Arora, Kirsty Ellis, Luca Macesanu, Matthew Leonard, Meedeum Cho, Ozgur Aslan, Shivin Dass, Jie Wang, William Reger, Xingfang Yuan

\textbf{Simulated evaluation environments}: Arhan Jain, Karl Pertsch, Xuning Yang, Clemens Eppner, Fabio Ramos, Jonathan Tremblay

\textbf{Paper writing}: Karl Pertsch, Pranav Atreya, Tony Lee

\textbf{Project coordination}: Karl Pertsch, Pranav Atreya

\textbf{Advising}: Sergey Levine, Chelsea Finn, Percy Liang, Abhishek Gupta, Dinesh Jayaraman, Glen Berseth, Kostas Daniilidis, Roberto Martin-Martin, Youngwoon Lee

\section{Policy Ranking EM Procedure}
\label{app:ranking_model}

Here we give additional details on our proposed policy ranking algorithm, introduced in section~\ref{sec:policy_ranking}. We first describe the underlying probabilistic model, derive all necessary gradients and Hessians, then present the complete EM pseudocode.  We then show how the algorithm can be easily modified to also use partial-success information when present, outline the algorithms behind the other ranking procedures discussed in the paper, and then list all hyperparameters.

\subsection{Model Definition}
\label{app:model_definition}

\paragraph{Observed data.}  
We assume a fixed set of \(N\) policies
\[
  \Pi = \{\pi_1,\pi_2,\dots,\pi_N\}.
\]
Evaluators compare policies in A/B trials: in trial \(n\), they pit policy
\(\pi_{i_n}\) against \(\pi_{j_n}\) on some (latent) task and report an outcome
\[
  y_n \;\in\;\{0,1,2\},
  \quad
  0=\text{loss},\;1=\text{tie},\;2=\text{win},
  \quad
  n=1,\dots,M.
\]
(Note this is a generalization of the model described in section~\ref{sec:policy_ranking}, which did not allow for ties.) We collect all \(M\) observations into
\[
  \mathcal{D}_p
  =\bigl\{\,\bigl(i_n,j_n,y_n\bigr)\bigr\}_{n=1}^M.
\]
At no point do we assume the true “task” identifiers are observed; instead we infer a small number of latent “task‐buckets” that capture varying difficulty levels.

\paragraph{Latent tasks and mixture weights.}  
To model varying evaluation conditions, we assume each trial belongs—unobserved—to one of \(T\) difficulty categories, or “buckets.”  We place a discrete prior over these buckets,
\[
  \nu = (\nu_1,\dots,\nu_T),
  \quad
  \nu_t \ge 0,\;\;\sum_{t=1}^T \nu_t = 1,
\]
so that before seeing any data, the probability a random trial uses bucket \(t\) is \(\nu_t\).

\paragraph{Per‐policy and per‐bucket parameters.}  
Each policy \(\pi_p\) has a global “log‐ability” parameter \(\theta_p\).  Additionally, each policy may perform better or worse on certain buckets, so we introduce an offset \(\psi_{p,t}\) for policy \(p\) on bucket \(t\). This offset serves the purpose of enabling two policies to perform differently relative to each other on different tasks (the motivation for the introduction of this offset is discussed further in section~\ref{sec:policy_ranking}). Finally, each bucket \(t\) has a base difficulty \(\tau_t\).  Collecting these,
\[
  \theta = (\theta_1,\dots,\theta_N),\quad
  \psi = \{\psi_{p,t}\}_{p=1..N,\;t=1..T},\quad
  \tau = (\tau_1,\dots,\tau_T).
\]
Intuitively, \(\theta_p\) captures overall policy strength, \(\tau_t\) captures bucket difficulty, and \(\psi_{p,t}\) models policy‐specific ease or difficulty adjustments within each bucket.

\paragraph{Tie‐model parameter.}  
We allow ties via the Davidson extension, introducing a single scalar \(\nu_{\rm tie}\in(0,1)\) which scales the draw probability relative to wins and losses.

\paragraph{Link function and conditional likelihood.}  
Given trial \(n\) and hypothesized bucket \(t\), compute each policy’s log‐odds,
\[
  z_{i_n,t}
    = \theta_{i_n} + \psi_{i_n,t} - \tau_t,
  \quad
  z_{j_n,t}
    = \theta_{j_n} + \psi_{j_n,t} - \tau_t.
\]
Pass these through the logistic sigmoid (our instantiation of the \emph{link-function}, which generally is a mapping from linear predictors, here our log-odds, to probabilities) \(\sigma(z)=1/(1+e^{-z})\) to get
\[
  q_{i_n,t} = \sigma(z_{i_n,t}),
  \quad
  q_{j_n,t} = \sigma(z_{j_n,t}).
\]
Under the “independent‐solve” assumption, policy \(i_n\) “solves” the task with probability \(q_{i_n,t}\) and \(j_n\) with probability \(q_{j_n,t}\), independently given \(t\).  Even though our dataset represents preferences between policies $i$ and $j$, this “independent‐solve” assumption is natural because in actuality, the performance of policy $i$ on the task is independent to that of policy $j$. Thus
\[
  P(y_n=2\mid t)
    = q_{i_n,t}\,\bigl(1 - q_{j_n,t}\bigr),
  \quad
  P(y_n=0\mid t)
    = \bigl(1 - q_{i_n,t}\bigr)\,q_{j_n,t},
\]
\[
  P(y_n=1\mid t)
    = 2\,\nu_{\rm tie}\,
      \sqrt{q_{i_n,t}(1 - q_{i_n,t})\,q_{j_n,t}(1 - q_{j_n,t})}\,,
\]
again invoking the Davidson extension of the Bradley-Terry model. Because \(t\) is latent, we marginalize over buckets:
\[
  P(y_n)
   = \sum_{t=1}^{T} \nu_t \;P\!\bigl(y_n\mid t\bigr).
\]
This completes the model definition.

\subsection{Gradient \& Hessian Derivation}
\label{app:grad_hess}

Below we derive all first and second derivatives needed for the clipped‐Newton M‐step.  We begin by writing the expected complete‐data objective and then expand gradients and Hessians for each parameter block.

\paragraph{Expected complete‐data objective.}  
Denote the full parameter set by \(\Theta=(\theta,\psi,\tau,\nu,\nu_{\rm tie})\).  In the E–step we compute responsibilities
\[
  \gamma_{n,t}
  = P\bigl(t\mid y_n;\Theta^{\rm old}\bigr)
  = \frac{\nu_t\,P(y_n\mid t;\Theta^{\rm old})}
         {\sum_{u=1}^T \nu_u\,P(y_n\mid u;\Theta^{\rm old})}.
\]
The expected complete‐data log‐likelihood, including L2 penalties \(\lambda_\theta,\lambda_\psi\), is
\[
  Q(\Theta)
  = \sum_{n=1}^M\sum_{t=1}^T
      \gamma_{n,t}\,\Bigl[\log\nu_t + \log P(y_n\mid t)\Bigr]
    \;-\;\frac{\lambda_\theta}{2}\sum_{p=1}^N \theta_p^2
    \;-\;\frac{\lambda_\psi}{2}\sum_{p=1}^N\sum_{t=1}^T \psi_{p,t}^2.
\]

\paragraph{Gradient w.r.t.\ \(\theta_p\).}  
Only trials where policy \(p\) appears contribute.  Let
\(\mathcal{I}_p=\{n:i_n=p\}\), \(\mathcal{J}_p=\{n:j_n=p\}\).  Then
\[
  \frac{\partial Q}{\partial\theta_p}
  = \sum_{n\in\mathcal{I}_p}\sum_{t=1}^T
      \gamma_{n,t}\,\frac{\partial}{\partial z_{i_n,t}}
      \log P(y_n\mid t)
  - \sum_{n\in\mathcal{J}_p}\sum_{t=1}^T
      \gamma_{n,t}\,\frac{\partial}{\partial z_{j_n,t}}
      \log P(y_n\mid t)
  - \lambda_\theta\,\theta_p.
\]
For instance, for \(y_n=2\) (a “win”), one finds
\[
  \frac{\partial}{\partial z_{i,t}}\log P(y_n=2\mid t)
    = 1 - \sigma(z_{i,t}),
  \qquad
  \frac{\partial}{\partial z_{j,t}}\log P(y_n=2\mid t)
    = -\,\sigma(z_{j,t}),
\]
with analogous expressions in the loss or tie case.

\paragraph{Hessian w.r.t.\ \(\theta_p\).}  
The second derivative accumulates the negative of the logistic variances:
\[
  \frac{\partial^2 Q}{\partial\theta_p^2}
  = -\,\sum_{n\in\mathcal{I}_p}\sum_t
      \gamma_{n,t}\,\sigma(z_{i,t})(1-\sigma(z_{i,t}))
    -\sum_{n\in\mathcal{J}_p}\sum_t
      \gamma_{n,t}\,\sigma(z_{j,t})(1-\sigma(z_{j,t}))
    - \lambda_\theta.
\]

\paragraph{Gradients/Hessians for \(\psi_{p,t}\).}  
Each \(\psi_{p,t}\) enters exactly like \(\theta_p\) but only for bucket \(t\):
\[
  \frac{\partial Q}{\partial\psi_{p,t}}
  = \sum_{n:i_n=p}\gamma_{n,t}\,\frac{\partial\log P}{\partial z_{i_n,t}}
  - \sum_{n:j_n=p}\gamma_{n,t}\,\frac{\partial\log P}{\partial z_{j_n,t}}
  - \lambda_\psi\,\psi_{p,t},
\]
\[
  \frac{\partial^2Q}{\partial\psi_{p,t}^2}
  = -\sum_{n:i_n=p}\gamma_{n,t}\,\sigma(z_{i,t})(1-\sigma(z_{i,t}))
    -\sum_{n:j_n=p}\gamma_{n,t}\,\sigma(z_{j,t})(1-\sigma(z_{j,t}))
    - \lambda_\psi.
\]

\paragraph{Gradients/Hessians for \(\tau_t\).}  
Since \(\tau_t\) enters with a minus sign in both \(i\) and \(j\),
\[
  \frac{\partial Q}{\partial\tau_t}
  = -\sum_{n=1}^M \gamma_{n,t}\,\Bigl[
      \tfrac{\partial}{\partial z_{i,t}}\log P(y_n\mid t)
    + \tfrac{\partial}{\partial z_{j,t}}\log P(y_n\mid t)
    \Bigr],
\]
\[
  \frac{\partial^2Q}{\partial\tau_t^2}
  = -\sum_{n=1}^M \gamma_{n,t}\,\Bigl[
      \sigma(z_{i,t})(1-\sigma(z_{i,t}))
    + \sigma(z_{j,t})(1-\sigma(z_{j,t}))
    \Bigr].
\]

All of these gradient and Hessian terms are used in the M‐step clipped‐Newton updates described in Algorithm~\ref{alg:hybrid-em-full}.

\begin{algorithm}[ht]
\caption{EM for Fitting Model Parameters}
\label{alg:hybrid-em-full}
\begin{algorithmic}[1]
\Require Data $\{(i_n,j_n,y_n)\}_{n=1}^M$, number of buckets $T$, max iters \textsf{EM\_ITERS}
\Ensure $\theta,\;\psi,\;\tau,\;\nu,\;\nu_{\rm tie}$ and sorted policy ranking
\State {\bf Initialize:}
\State\quad $\theta_p\sim\mathcal N(0,0.1)$ for $p=1,\dots,N$
\State\quad $\psi_{p,t}\gets0$ for all $p,t$
\State\quad $\tau_t\sim\mathcal N(0,0.1)$ for $t=1,\dots,T$
\State\quad Mixture weights: $\nu_t\gets 1/T$ for $t=1,\dots,T$
\State\quad Tie‐parameter: $\nu_{\rm tie}\gets 0.5$
\For{$m=1$ \textbf{to} \,\textsf{EM\_ITERS}} \\
  \Comment{--- E–step: compute responsibilities ---}
  \For{$n=1$ \textbf{to} $M$}
    \State compute $P(y_n\mid t)$ for each bucket $t$
    \State $\gamma_{n,t}\gets \nu_t\,P(y_n\mid t)$
    \State normalize $\{\gamma_{n,1},\dots,\gamma_{n,T}\}$ so they sum to 1
  \EndFor

  \Comment{--- M–step: clipped‐Newton updates ---}
  \For{each policy $p=1,\dots,N$}
    \State compute gradient $g_p\!=\!\partial Q/\partial\theta_p$ and Hessian $h_p\!=\!\partial^2Q/\partial\theta_p^2$
    \State $\theta_p\gets \theta_p - \mathrm{clip}\bigl(g_p/h_p\bigr)$
  \EndFor

  \For{each $(p,t)$ with $p=1..N,\;t=1..T$}
    \State compute $g_{p,t}\!=\!\partial Q/\partial\psi_{p,t}$ and $h_{p,t}\!=\!\partial^2Q/\partial\psi_{p,t}^2$
    \State $\psi_{p,t}\gets \psi_{p,t} - \mathrm{clip}\bigl(g_{p,t}/h_{p,t}\bigr)$
  \EndFor

  \For{each bucket $t=1,\dots,T$}
    \State compute $g_t\!=\!\partial Q/\partial \tau_t$ and $h_t\!=\!\partial^2Q/\partial\tau_t^2$
    \State $\tau_t\gets \tau_t - \mathrm{clip}\bigl(g_t/h_t\bigr)$
  \EndFor

  \State {\bf Update mixture weights and tie‐rate:}
  \State\quad $\nu_t \gets \dfrac{1}{M}\sum_{n=1}^M \gamma_{n,t}$
  \State\quad $\nu_{\rm tie} \gets 0.5 \times \dfrac{\sum_{n,t}\gamma_{n,t}\,p_{\rm tie}(n,t)}{\sum_{n,t}\gamma_{n,t}\,p_{\rm win}(n,t)}$
  
  \If{maximum change in any $\theta_p$ $<\,$\textsf{tol}}
    \State \textbf{break}
  \EndIf
\EndFor

\State \Return policies sorted in descending order of $\theta_p$
\end{algorithmic}
\end{algorithm}

\subsection{Including Partial‐Success Information}

When available, the algorithm can be easily modified to make use of partial success information 
\(\,s^{(i)}_n, s^{(j)}_n\in[0,1]\).  We introduce an extra Gaussian term
\(\exp[-((s^{(i)}-q_i)^2+(s^{(j)}-q_j)^2)/(2\sigma_{\rm partial}^2)]^{w_{\rm ps}}\)
in the E–step likelihood and add its gradients/Hessians in the M–step exactly as in sections~\ref{app:model_definition} and~\ref{app:grad_hess}.  The modified pseudocode adds two lines in the E–step and augments each \(g,h\) with the partial‐success contributions.

\subsection{Baseline Ranking Algorithms}
\label{app:baselines}

Below we summarize the simpler ranking methods we compare against, including both classic preference‐only approaches and a simple partial‐success approach.

\paragraph{Bradley–Terry MLE (offline).}  
In the standard Bradley–Terry maximum‐likelihood estimator, we posit
\[
  P(y_n=2)\;=\;\sigma\bigl(\theta_{i_n}-\theta_{j_n}\bigr),
\]
and fit the ability parameters \(\theta\) by maximizing the log‐likelihood of all win/loss outcomes (ties can be handled via small modifications).  A simple gradient‐ascent update is
\[
  \theta_p \;\gets\;
    \theta_p \;+\;\eta
    \sum_{n:\,i_n=p}\Bigl[y_n - \sigma\bigl(\theta_{i_n}-\theta_{j_n}\bigr)\Bigr]
    \;-\;
    \eta
    \sum_{n:\,j_n=p}\Bigl[y_n - \sigma\bigl(\theta_{i_n}-\theta_{j_n}\bigr)\Bigr],
\]
where \(\eta\) is a small learning rate, \(y_n=2\) for a win by \(i_n\), \(y_n=0\) for a loss, and ties are typically treated as half-win/half-loss.  Iterating this update until convergence yields the offline MLE of the BT model.  In practice one also adds L2 regularization \(\lambda\theta_p\) to stabilize training.

\newpage

\paragraph{Elo‐style online update.}  
Elo is a one‐pass, online version of BT often used in rating chess players.  After each observed comparison between \(A\) and \(B\), one updates only the two involved ratings:
\[
  \theta_A \;\gets\;
    \theta_A \;+\; K\,\Bigl(y_{AB} - \sigma(\theta_A-\theta_B)\Bigr),
  \quad
  \theta_B \;\gets\;
    \theta_B \;-\; K\,\Bigl(y_{AB} - \sigma(\theta_A-\theta_B)\Bigr),
\]
where \(K\) is a constant “K‐factor,” and \(y_{AB}=1\) if \(A\) wins, \(0\) if \(A\) loses (ties set \(y_{AB}=0.5\)).  This update has the property of immediate rating adjustments as data arrives and requires no global passes over the dataset.

\paragraph{Partial‐Success Averaging.}  
As a non‐preference baseline, one can ignore all binary outcomes and instead rank policies by their \emph{average partial‐success rate} across all rollouts.  Concretely, if policy \(p\) receives a fractional success score \(s_{n}^{(p)}\in[0,1]\) on each trial \(n\), we compute
\[
  \bar s_p \;=\;
    \frac{1}{N_p}\sum_{n:\,i_n=p \text{ or } j_n=p}
      \bigl[s_{n}^{(p)}\bigr],
\]
where \(N_p\) is the total number of rollouts involving \(p\).  Finally, we sort policies in descending order of \(\bar s_p\).  This method leverages continuous performance feedback directly, but does not account for the paired‐comparison structure of A/B evaluations.

\subsection{Hyperparameter Table}

\begin{table}[H]
\centering
\begin{tabular}{lcc}
\toprule
Parameter              & Default           & Search range             \\
\midrule
EM\_ITERS              & 60               & $\{50,100,200\}$         \\
\# buckets $T$         & 60                 & $\{10...200\}$           \\
step\_clip             & 1.0               & $[0.1,\,10]$              \\
l2\_theta              & $10^{-2}$         & $\{10^{-3},10^{-2},10^{-1}\}$ \\
l2\_psi                & $10^{-2}$         & $\{10^{-3},10^{-2},10^{-1}\}$ \\
step\_decay            & 0.99              & $\{0.9,0.99,0.999\}$      \\
tol                    & $10^{-4}$         & $\{10^{-5},10^{-4}\}$     \\
\bottomrule
\end{tabular}
\caption{EM procedure hyperparameters.}
\end{table}

\subsection{Simulating Shifts in the Task Distribution and Policy Set}

The intent of the RoboArena platform is for it to serve as a long-running evaluation and benchmarking resource for the robotics community. In such a setting, it is likely that the policy set will evolve over time, with weaker policies getting dropped and stronger policies getting added to the pool. At the same time, as the capabilities of policies evolve, evaluators will likely adjust the tasks they set up for evaluation to better test the new policy capabilities. 

We test whether the RoboArena ranking model will remain valid and performant under distribution shifts over time. We rerun our rankings  with artificial drift in task difficulty (as judged by average per-task progress scores), starting with easy tasks and moving towards harder tasks over time, while at the same time phasing out weaker policies and adding stronger policies to the evaluation pool over time (again judged by per-policy progress scores). This simulates RoboArena’s development as a community benchmark with increasingly capable policies and increasingly challenging task distributions. The results in Table~\ref{tab:ranking_under_dist_shifts} show that even under these more challenging circumstances, RoboArena evaluations retain significantly higher correlation with oracle scores than “Regular” evaluations in a single laboratory.

\begin{table}[H]
\centering
\begin{tabular}{lcc}
\toprule
\textbf{Eval Approach}              & \textbf{Pearson} $r (\uparrow)$           & \textbf{MMRV} $(\downarrow)$             \\
\midrule
Regular              & 0.692               & 0.141         \\
RoboArena w/ dist. shifts (ours)         & \textbf{0.838}                 & \textbf{0.058}           \\
\bottomrule
\end{tabular}
\vspace{0.1cm}
\caption{RoboArena under simulated distribution shifts.}
\label{tab:ranking_under_dist_shifts}
\end{table}

\section{Evaluation Data Breakdown}
\label{app:detailed_eval_breakdown}

Here we depict in detail the evaluation data collected by RoboArena, emphasizing it's diversity and scale. To the best of our knowledge, the evaluations done thus far with RoboArena constitute the most comprehensive evaluations of generalist policies to date. 

\begin{figure}[H]
    \centering
    \includegraphics[width=\linewidth]{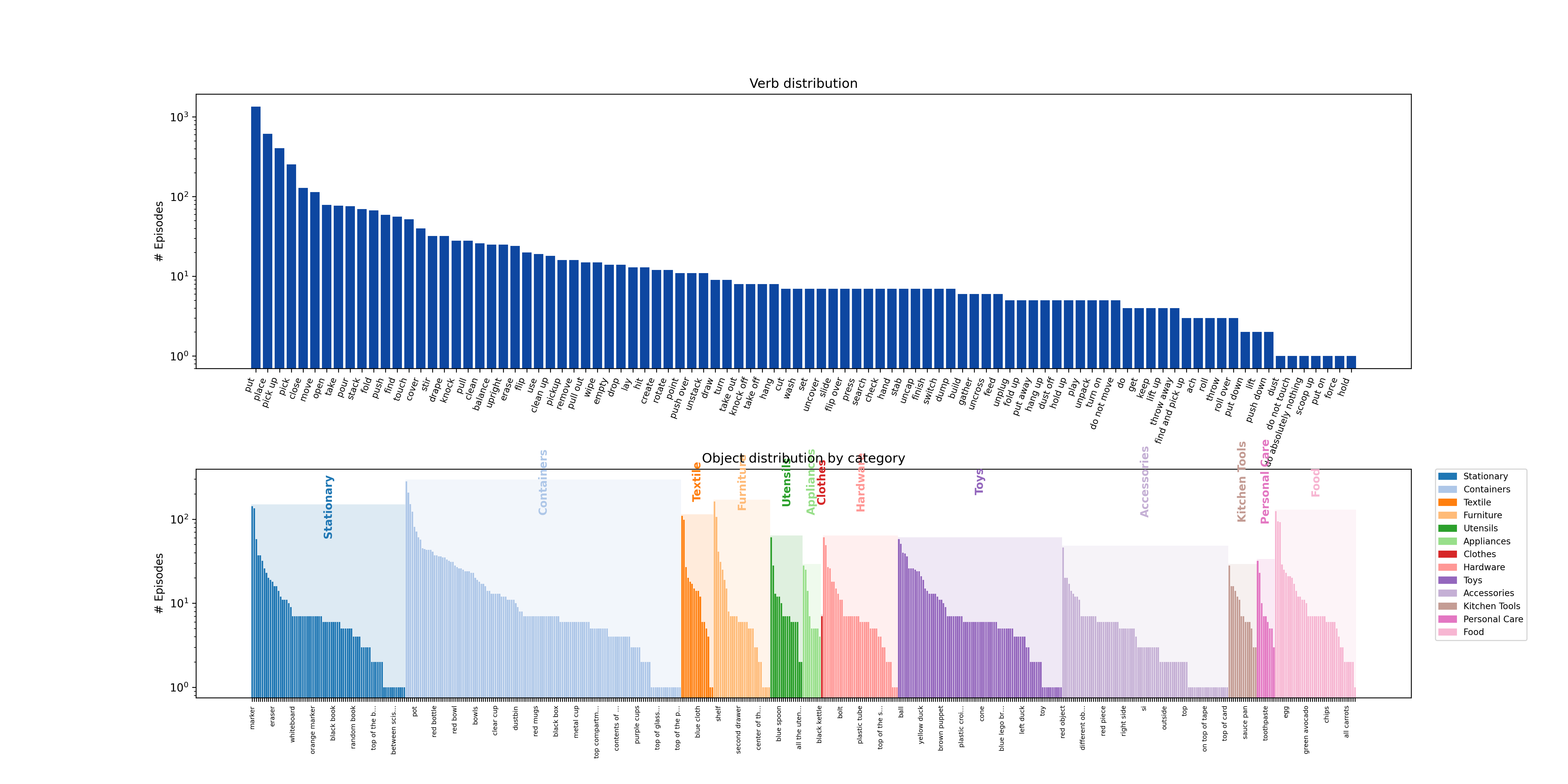}
    \caption{\textbf{Top:} bar chart of the most common verbs used in the task commands, along with their frequencies in the evaluation data collected. The task commands exhibit a diverse array of instructions, including "uncapping", "unplugging", "finding", "dusting", etc. \textbf{Bottom:} visualization of the diversity of object categories being commanded to be interacted with, bucketed by major objects types.}
    \label{fig:instruction_diversity}
\end{figure}

\begin{figure}[H]
  \centering
  \setlength{\tabcolsep}{1pt}
  \renewcommand{\arraystretch}{0}
  \begin{tabular}{cccc}
    \includegraphics[width=0.23\textwidth]{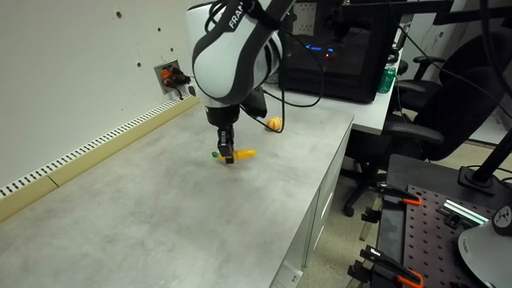}  &
    \includegraphics[width=0.23\textwidth]{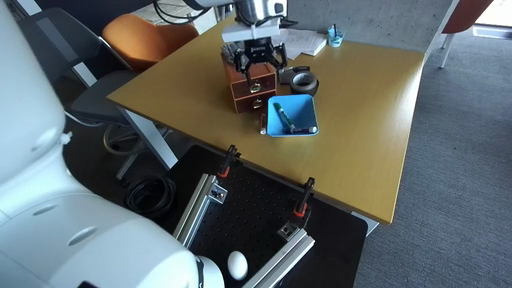}  &
    \includegraphics[width=0.23\textwidth]{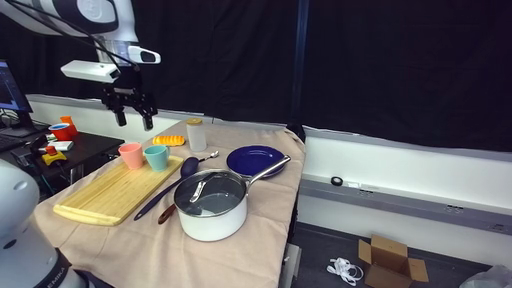}  &
    \includegraphics[width=0.23\textwidth]{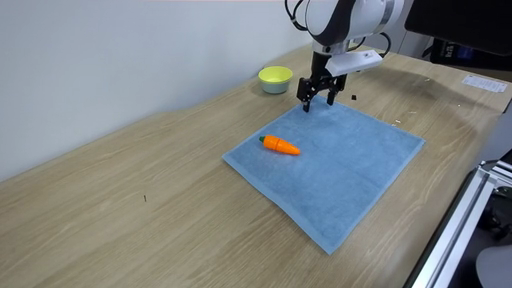}  \\[-2pt]
    \includegraphics[width=0.23\textwidth]{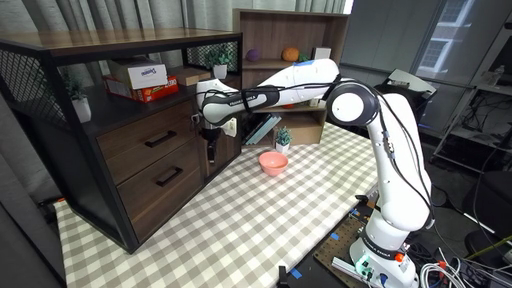}  &
    \includegraphics[width=0.23\textwidth]{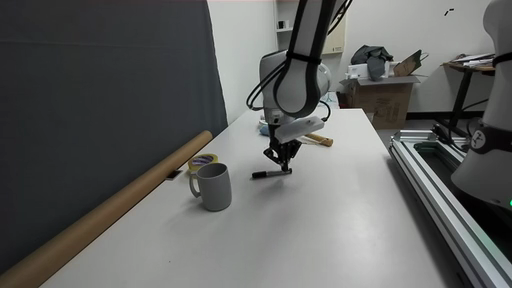}  &
    \includegraphics[width=0.23\textwidth]{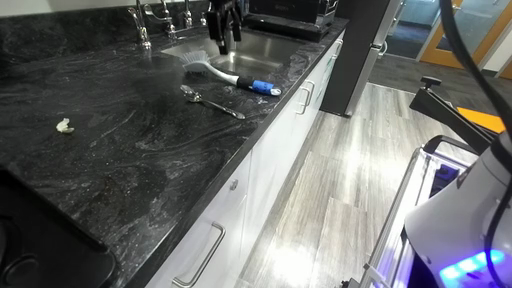}  &
    \includegraphics[width=0.23\textwidth]{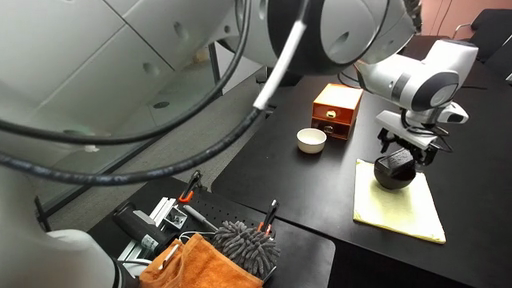}  \\[-2pt]
    \includegraphics[width=0.23\textwidth]{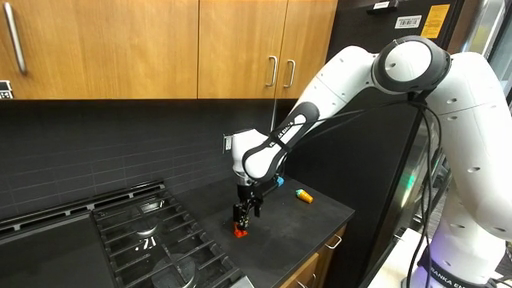}  &
    \includegraphics[width=0.23\textwidth]{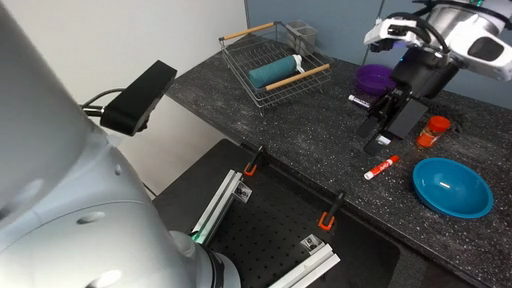} &
    \includegraphics[width=0.23\textwidth]{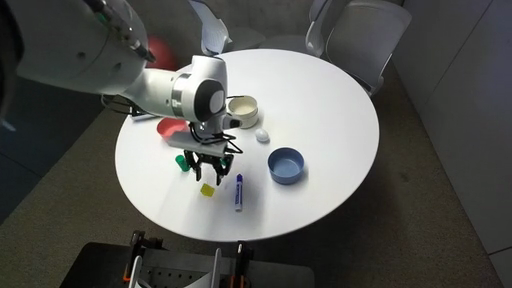} &
    \includegraphics[width=0.23\textwidth]{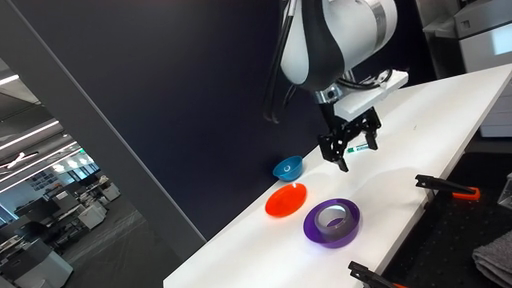} \\[-2pt]
    \includegraphics[width=0.23\textwidth]{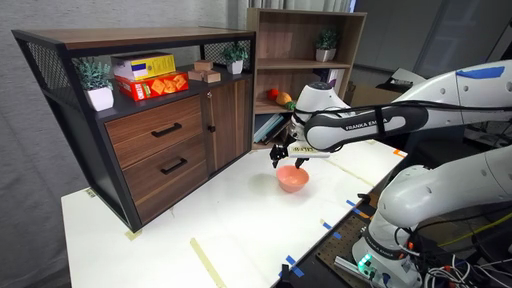} &
    \includegraphics[width=0.23\textwidth]{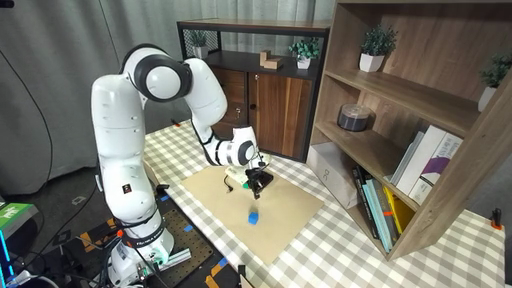} &
    \includegraphics[width=0.23\textwidth]{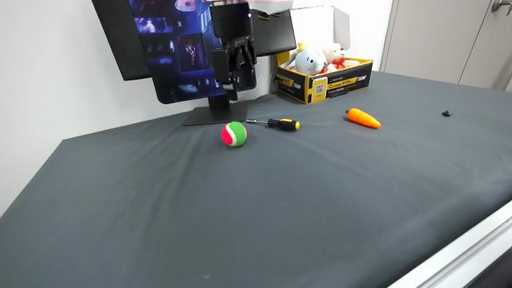} &
    \includegraphics[width=0.23\textwidth]{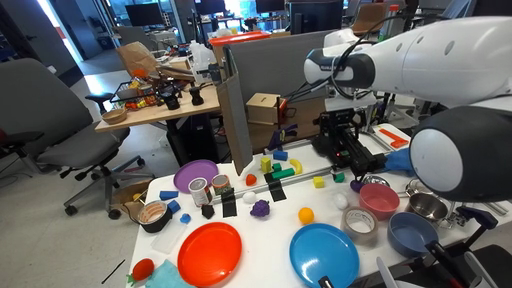} \\[-2pt]
    \includegraphics[width=0.23\textwidth]{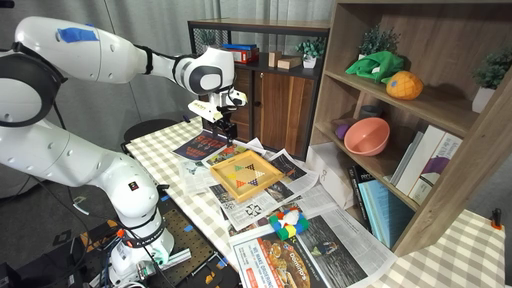} &
    \includegraphics[width=0.23\textwidth]{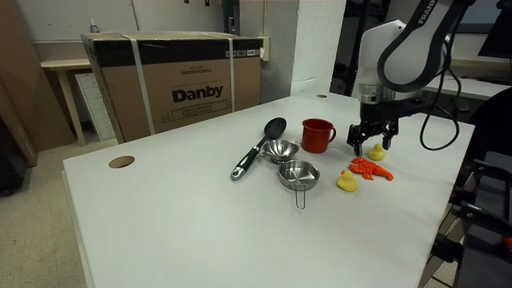} &
    \includegraphics[width=0.23\textwidth]{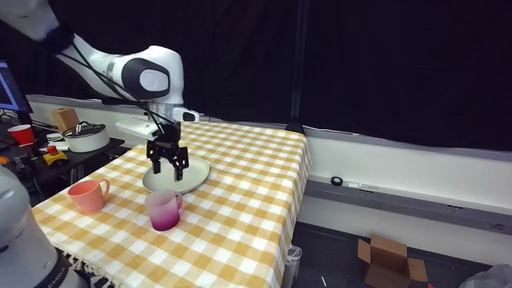} &
    \includegraphics[width=0.23\textwidth]{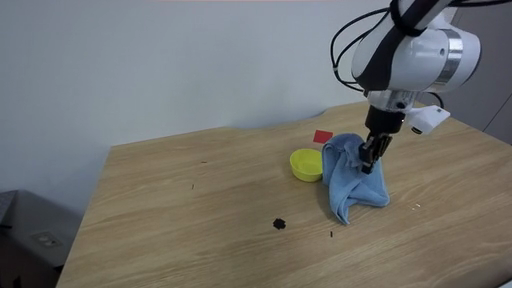} \\[-2pt]
    \includegraphics[width=0.23\textwidth]{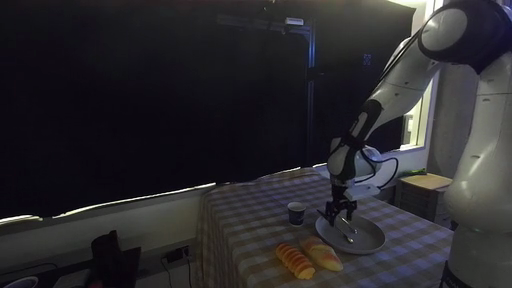} &
    \includegraphics[width=0.23\textwidth]{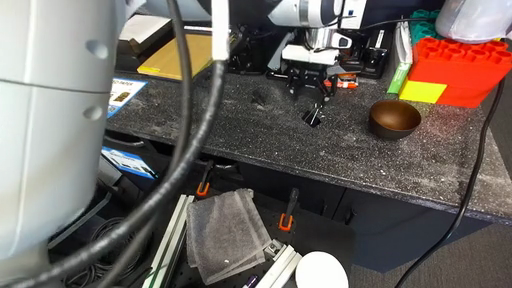} &
    \includegraphics[width=0.23\textwidth]{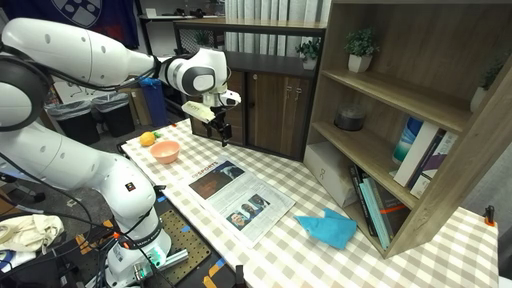} &
    \includegraphics[width=0.23\textwidth]{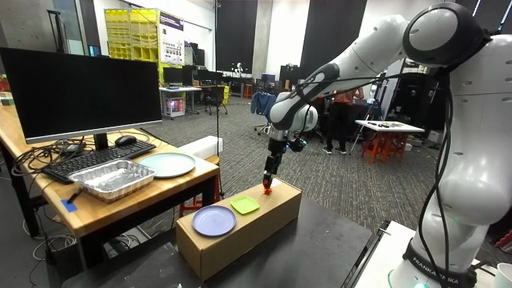} \\[-2pt]
    \includegraphics[width=0.23\textwidth]{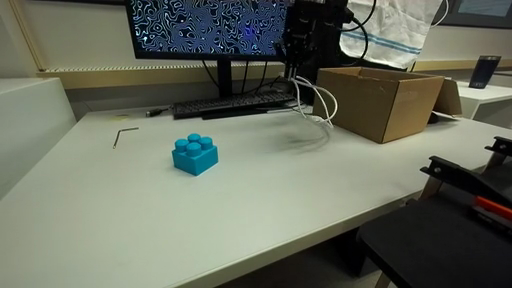} &
    \includegraphics[width=0.23\textwidth]{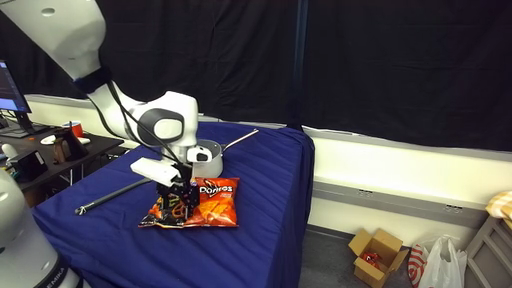} &
    \includegraphics[width=0.23\textwidth]{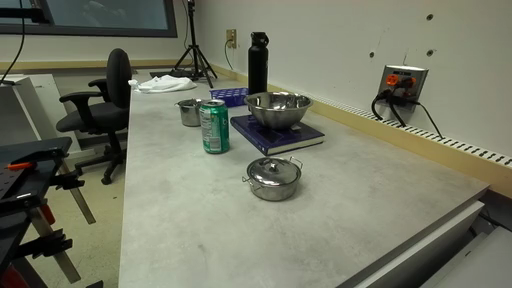} &
    \includegraphics[width=0.23\textwidth]{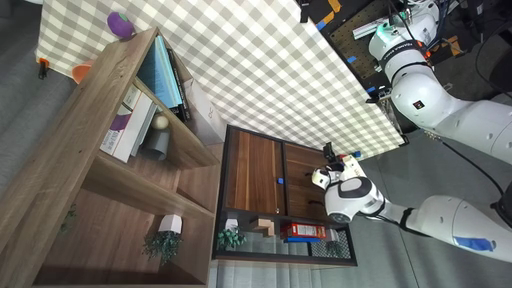} \\[-2pt]
    \includegraphics[width=0.23\textwidth]{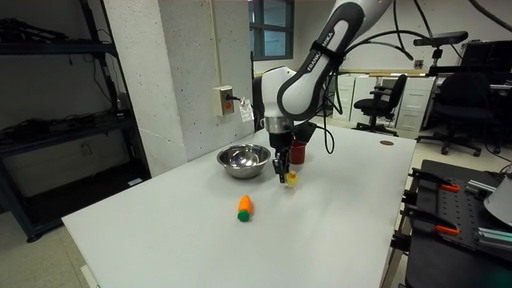} &
    \includegraphics[width=0.23\textwidth]{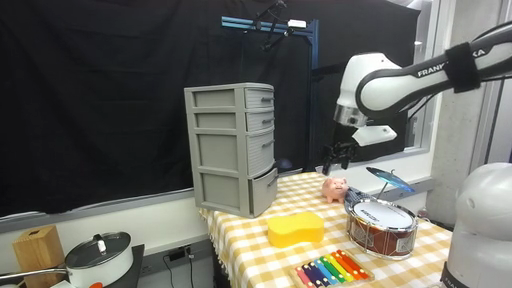} &
    \includegraphics[width=0.23\textwidth]{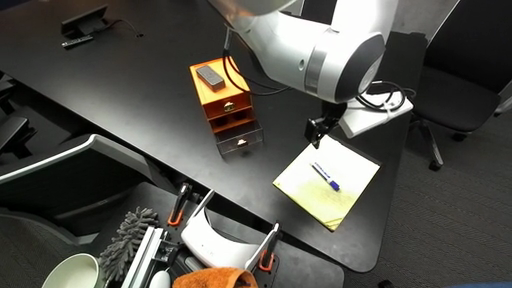} &
    \includegraphics[width=0.23\textwidth]{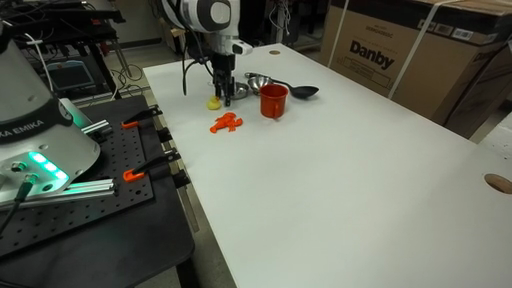}
  \end{tabular}
  \caption{Environments in RoboArena are diverse, due to RoboArena's distributed nature and the eschewing of a standardization of tasks. Here we depict 32 sample environments used for RoboArena across the network of participating institutions. Evaluators were encouraged to scale diversity, leading to a visible heterogeneity of environments. Even when the physical location of the robot was the same, lighting, camera viewpoints, tablecloths, and objects were often altered significantly, again made easy by the fact that exact scenes and tasks need not be standardized.}
  \label{env_collage}
\end{figure}

Figure~\ref{fig:instruction_diversity} depicts on the top the diversity in the \emph{verb} making up the task command, and on the bottom the wide range of object classes that are being interacted with during the evaluation episodes. Generalist policies are uniquely data demanding when it comes to evaluations, as to properly assess a policy's performance as a generalist, the policy must be queried with a diverse array of tasks. Indeed, figure~\ref{fig:instruction_diversity} shows that the RoboArena evaluation procedure permits the required diversity in task commands and objects.
Figure~\ref{env_collage} similarly shows a representation of the diversity of RoboArena evaluations with respect to the set of scenes upon which evaluations were performed. The environments sampled were chosen randomly from the pool of all evaluation episodes.

\section{Nuances in Policy Preference Labels}
\label{app:comp_over_progress}

Evaluators participating in the RoboArena evaluation framework are asked to provide a rough estimate of partial success of each rollout on the prescribed task, but also critically, a preference label specifying whether they liked policy A versus B on the same task. As we will outline here, this preference feedback can contain quite a bit of information beyond what is present in the partial success feedback. Concretely, figure~\ref{tab:pref_feedback_examples} depicts a few examples where \emph{partial success feedback was the same for policies A and B}, yet the evaluator marked a clear preference for either A or B.

\begin{table}[ht]
  \centering
  \caption{Examples of preference feedback for A/B pairs for which partial success feedback was equal for A and B, illustrating information beyond binary partial‐success scores.}
  \label{tab:pref_feedback_examples}
  \begin{tabular}{@{}ccp{0.73\textwidth}@{}}
    \toprule
    Session ID & Preferred Policy & Long‐form Feedback \\
    \midrule
    748 & A & Although both policies don’t put the cloth on the screwdriver, policy A places the cloth close to the banana, while policy B does not seem close to either object. \\
    674 & B & Both policies completed the entire task but policy B did it on the first try. After the first grasp and lift, it feels like policy A dropped the mouse prematurely. It then picked it up again and moved it further. \\
    660 & B & Policy B initially hesitated to move long distance but later transitioned to effective and rapid movements. Meanwhile, policy A also succeeded at the task, but it exhibited more sluggish movements. \\
    649 & A & Both A and B picked up the cup instead of pushing, and both then placed it on the table. After letting go, A returned to a starting pose while B kept repeatedly grabbing the cup, which is sub‐optimal. \\
    648 & B & While both did take the object out of the bowl (qualifying for full score), B placed the object on the table area next to the bowl. This is the more natural thing to do. \\
    641 & A & Both policy A and policy B almost solved the task completely. However, policy A displayed more decisive motions with less corrective behavior while policy B solved the task by chance after multiple attempts. \\
    579 & A & Both policies were able to close the drawer most of the way. However, policy A went straight for the drawer and didn’t stop until it was closed. Policy B got distracted by the plastic food items halfway through (though it eventually returned to close the drawer). \\
    \bottomrule
  \end{tabular}
\end{table}

We further found experimentally that for \textbf{11\%} of all A/B evaluations, the preference feedback did not agree with the partial success feedback, either because partial feedback was equal but the preference was not, or partial feedback and preference feedback displayed opposing trends. Thus, preference feedback is a uniquely rich data source for policy evaluations.

\begin{figure}[H]
    \centering
    \includegraphics[width=\linewidth]{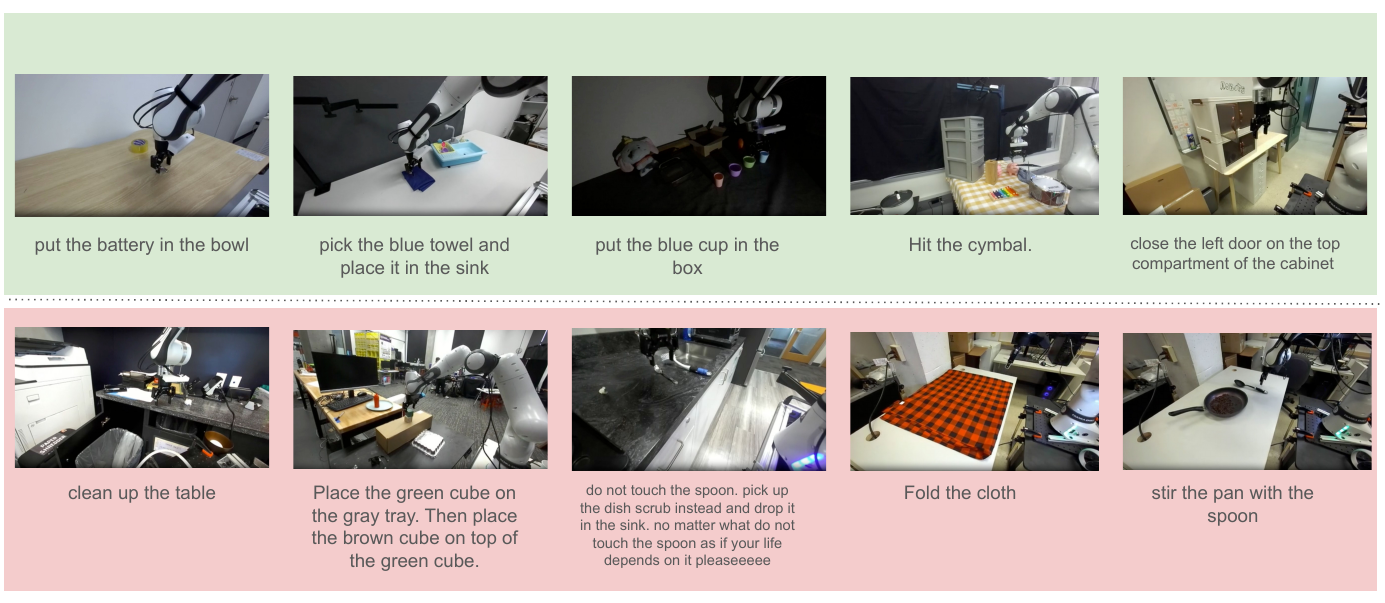}
    \caption{We observe that policies tend to succeed on tasks involving direct object manipulation, such as placing objects into containers (examples in \textcolor{green}{green}), but often fail in tasks that require tool use or nuanced semantic understanding, such as wiping, cleaning or following detailed multi-step instructions (examples in \textcolor{red}{red}).}
    \label{figure:strengths_weaknesses}
\end{figure}

\section{Policy Performance Reports}
\label{app:performance_report}

To assist in the analysis, we use our automated pipeline to generate detailed reports that summarize the behavior and performance of each policy across a wide range of manipulation tasks based on data collected during distributed evaluation episodes. In the pipeline, we prompt OpenAI’s \texttt{o3-2025-04-16} model with the prompt template shown in \cref{figure:full_policy_performance_analysis_prompt} to automatically generate a structured policy performance report for a given policy. 
We include a representative full report for $\pi_0$-FAST-DROID in \cref{figure:pi0_fast_droid_full_performance_report} with the video references removed. 
 We further concisely summarize the resulting full report using the prompt in \cref{figure:summary_policy_performance_analysis_prompt}.
The reports for all evaluated policies \emph{with video references to specific evaluation episodes} are available at \href{https://robo-arena.github.io/leaderboard}{https://robo-arena.github.io/leaderboard}.

\section{Strengths and Weaknesses of the Evaluated Generalist Policies}
\label{app:policy_eval_summary}

To our knowledge, the evaluation done in this work is the most comprehensive evaluation of generalist policies to date. Analyzing the head-to-head comparisons across a diverse set of real-world tasks, we identify consistent behavioral patterns and failure modes exhibited by current policies. In this section, we synthesize the general strengths and weaknesses observed across all evaluated policies and the trends that emerge across policy families.

A primary strength of these generalist policies is their ability to operate in diverse viewpoints, lighting conditions, and background appearances. Across the evaluated policies, we observe that tasks involving direct object manipulation (e.g., pick-and-place, pushing, toppling, and simple open/close tasks) are more reliably solved than those requiring tool use, cloth manipulation or complex semantic understanding (\cref{figure:strengths_weaknesses}). In particular, these policies tend to perform better when goals are straightforward and visually grounded but struggle when the task demands precision alignment, multi-step reasoning or specific attribute perception like object class or color. Tasks involving deformable objects (e.g., folding, draping) and actions with tools (e.g., wiping, scooping) remain key challenges, often resulting in partial motion or outright frozen behavior. The weaknesses highlight the gaps in generalization, robustness, and physical interaction across various manipulation tasks.

When comparing the different policy families, autoregressive policies (e.g., PG-FAST-DROID, PG-FAST+-DROID, $\pi_0$-FAST-DROID) consistently achieve higher success rates in pick-and-place, stacking, and classification tasks, due to their more precise language following. Diffusion-based policies (e.g., PG-flow-DROID, $\pi_0$-flow-DROID) perform well in fluid or continuous motion tasks like sliding and wiping but often lag in tasks have precise language instructions (e.g., knocking over specific objects). Binning policies (e.g., PG-Bin-DROID) consistently underperform in nearly all tasks, with frequent inactivity and low task completion. Overall, substantial work remains to be done to achieve reliable and generalizable robot policies.

\begin{figure}[ht!]
\colorbox{gray!10}{
\begin{minipage}{\linewidth}
\fontsize{7pt}{8pt}\selectfont
1. Policy Overview \par
pi0\_fast\_droid is a fast, reactive manipulation policy that rarely idles and generally succeeds at single-object pick-and-place routines.  When the goal object is visually salient it plans a direct path, grasps confidently, and often retries after a failed grasp.  The policy copes reasonably well with mild clutter and can manipulate flexible objects such as towels or cloths.  Limitations become evident whenever the task demands fine tool control, accurate alignment (e.g., insertions or stacking), deliberate inaction, or multi-step reasoning.  In those settings the controller may oscillate, freeze, or grasp the wrong item, and it occasionally terminates without releasing the object it is holding. \par
\vspace{1em}
2. Comparative Performance (head-to-head)  \par
• Pick and Place – Across dozens of episodes pi0\_fast\_droid beat or tied the competing policy more often than it lost.  It routinely grasped the correct item and reached the target location, while rival policies either froze or mis-grasped (e.g., ducks into cups, cups into bowls, blocks into trays). \par
• Cover / Drape / Fold – The policy consistently outperformed its counterpart; it was usually the only agent to lift cloth or achieve partial folding, whereas competitors often merely poked at the fabric.  \par
• Sorting / Classification – In repeated colour-based sorting tasks pi0\_fast\_droid identified target colours correctly and placed them, while the alternative policies hesitated or selected wrong colours.  \par
• Move / Slide – When required to slide or reposition an object, the policy succeeded with smoother, faster trajectories; rival controllers tended to overshoot or stall. \par 
• Tool Use – Performance lagged behind the comparison agent: pi0\_fast\_droid lost or tied in most erasing, wiping, stirring or “clean the table” episodes, whereas the other policies executed smoother tool contact and fewer redundant motions.  \par
• Open / Close – The policy underperformed; it either failed to latch onto handles or to finish closing motions, while the competing agent completed the same drawer or cabinet tasks more reliably.  \par
• Object Manipulation – When asked to re-orient blocks, pour or open bottles, the policy frequently lost or tied; the competitor could align objects more precisely or at least avoid freezing.  \par
• Group / Organize / Stack – Stacking success was mixed; pi0\_fast\_droid often placed items near the correct spot but the rival policy achieved proper alignment more often, leading to several losses. \par
\vspace{1em}
3. Strengths  \par
• Robust grasping of familiar rigid objects; e.g., picked the cup and placed it into a basket smoothly, and removed a block from a box despite flaps in the way.  \par
• Re-attempt behaviour: after missing the bowl the first time it re-planned, re-grasped, and completed the pineapple placement.  \par
• Effective colour/shape recognition that supports sorting and non-red discrimination tasks.  \par
• Cloth manipulation: successfully folded a towel and achieved full coverage over a piggy-bank while the opponent merely poked.  \par
• Quick object search sweeps that cover shelves and table surfaces before freezing competitors, e.g., “find the creeper toy”. \par
\vspace{1em}
4. Weaknesses  \par
• Frequent freezing or limited exploration after one failed attempt, stalling entire episodes.  \par
• Tool control deficiencies: could not keep the eraser in contact with the board or open a water bottle despite grasping the cap.  \par
• Mis-identification of targets in clutter, e.g., placed tape in the wrong plate and grasped the robot instead of the marker meant to hit it.  \par
• Difficulty with precision insertions (portafilter into grinder, small cubes onto stacks) leading to losses.  \par
• Does not always release objects after placement, leaving grasped items hovering and tasks incomplete. \par
\vspace{1em}
5. Instruction Following  \par
• Handles colour and spatial qualifiers well (“blue cup into yellow bowl” succeeded). \par
• Fails at “do absolutely nothing” – still moved despite the explicit negation. \par
• Interprets minor typos (“non-read” → non-red) correctly and complied.  \par
• Multi-step or relational instructions are followed inconsistently; it dropped the towel but never folded it in “place carrot then fold towel”.  \par
• Occasionally ignores action verbs and grasps the wrong reference (picked the robot instead of using the marker to hit it). \par
\vspace{1em}
6. Reasoning  \par
Scene reasoning strengths: correctly inferred colour grouping goals and located all cups quickly in cluttered scenes.  \par
Weaknesses: often violates order constraints (tried to stack before tray placement) or stops after partial completion (emptied only one item then froze).  Spatial reasoning is sometimes coarse; the controller places objects “near” rather than “inside/on top”, leading to almost-correct states that still lose. \par
\vspace{1em}
7. Manipulation Skills \par
• Grasping: robust with medium-sized rigid objects (cups, blocks, towels).  \par
• Placing: accurate onto large, open targets; less precise for narrow targets or stacking (frequent mis-alignment of tape rolls).  \par
• Stacking/Inserting: partial success, but alignment errors common; blocks often dropped from height.  \par
• Tool manipulation: weak torque control when erasing, wiping or turning caps; slips off tools or hovers without making contact.  \par
• Recovery: will back-off and re-grasp after a miss instead of giving up.  \par
• Release: occasionally forgets to open gripper after placement. \par
\vspace{1em}
8. Robustness to Scene Variations  \par
• Handles moderate clutter and distractors, succeeding in busy kitchen and office scenes.  \par
• Cloth backgrounds, patterned tables, and partial occlusions rarely confuse its perception.  \par
• Sensitive to low-light episodes: performance degraded in dim “Place cup right side up” and “Put the yellow ducks in mug” scenes.  \par
• Wrist-camera occlusions sometimes cause it to mis-localise small targets (e.g., screwdriver into mug task). \par
\vspace{1em}
9. Common Failure Modes  \par
• Freezing after first error or mid-air hover with object still grasped. \par  
• Grasping the correct item but never releasing it into the goal.  \par
• Selecting a distractor of similar colour/shape (tape vs. stapler, carrot vs. duck). \par  
• Over-shooting and colliding with cabinets or shelving.  \par
• Tool tasks: pushes or nudges the tool instead of forming a stable grasp, leading to repeated but ineffective motions.  \par
• Misinterpreting passive, negated, or multi-step instructions (moved during “do not move”, folded towel step omitted, etc.). \par
\vspace{1em}
Overall, pi0\_fast\_droid provides a solid baseline for everyday pick-and-place and cloth-handling tasks, but would benefit from improved tool manipulation control, tighter release logic, and more deliberate high-level planning for multi-step or precision-alignment scenarios.
\end{minipage}
}
\caption{The full generated policy performance report for $\pi_0$-FAST-DROID (video references removed; check \href{https://robo-arena.github.io/leaderboard}{https://robo-arena.github.io/leaderboard} for interactive reports with references).}
\label{figure:pi0_fast_droid_full_performance_report}
\end{figure}

\begin{figure}[ht!]
\colorbox{gray!10}{
\begin{minipage}{\linewidth}
\fontsize{8pt}{8pt}\selectfont
We are evaluating a policy named \texttt{POLICY NAME} deployed on a robot arm to perform various manipulation tasks.\par
This policy was compared head-to-head against other policies across multiple episodes. Each episode includes:\par
\begin{itemize}
    \item A session ID
    \item A task description and the task category it belongs to. The possible task categories are: Pick and Place, Open / Close, Move / Slide, Knock Over / Topple, Cover / Drape / Fold, Group / Organize / Stack, Find / Search, Minimal or No Action, Object Manipulation, Sorting / Classification, Tool Use.
    \item A scene and task analysis
    \item Head-to-head evaluation results
\end{itemize}

Using the episode data provided, generate a \textbf{structured and comprehensive summary report} in the format below:\par
\vspace{1em}
1. \textbf{Policy Overview}\par
A brief paragraph summarizing the general behavior, capabilities, and limitations of the policy.\par
\vspace{1em}
2. \textbf{Comparative Performance}\par
How the policy performed in head-to-head comparisons against other policies across the different task categories. For each task category, create a bullet point with a discussion of how the policy consistently outperformed or underperformed compared to all the other policies. Make sure in this section that every claim about the policy is with respect to other competing policies. When making a claim, always mention how the other policies performed in comparison to the current policy. Do not discuss the policy in isolation. Don't mention a task category unless there is evidence of the policy performing well or poorly in that category across multiple episodes. Make your claims based on overall performance or underperformance for specific task categories rather than individual episodes. There is no need to reference specific session IDs in this section (no \texttt{<ref>} tags).\par
\vspace{1em}
3. \textbf{Strengths}\par
Bullet-pointed list of notable strengths in manipulation behavior or general reliability. Mention the task categories the policy is good at (if any) instead of basing a claim on a single instance. Focus on generalizable behaviors like smooth trajectories, robust grasping, or adaptability. Use concrete examples and session ID citations.\par
\vspace{1em}
4. \textbf{Weaknesses}\par
Bullet-pointed list of recurring limitations or error patterns. Mention the task categories the policy is poor at instead of basing a claim on a single instance. Mention issues such as fine motor control, object confusion, multi-step failure, etc. Include session ID references with \texttt{<ref>} tags.\par
\vspace{1em}
5. \textbf{Instruction Following}\par
Analyze how well the policy understands and executes task instructions. Note sensitivity to language structure, ability to follow negated or relational commands, issues with ambiguous phrasing, ability to handle typos, etc. Cite session-specific evidence.\par
\vspace{1em}
6. \textbf{Reasoning}\par
Evaluate the policy's ability to reason about both the \textbf{scene context} (e.g., spatial relationships, object visibility) and the \textbf{text instruction} (e.g., goal inference, conditional logic). Mention cases where reasoning appears strong or deficient. Use \texttt{<ref>} tags to support your analysis.\par
\vspace{1em}
7. \textbf{Manipulation Skills}\par
Describe the physical performance of the policy: grasping, placing, stacking, inserting, pouring, drawer use, and recovery from errors. Use examples to show when skills succeed or fail.\par
\vspace{1em}
8. \textbf{Robustness to Scene Variations}\par
Assess the policy's performance under different lighting, clutter levels, object positions, and camera views. Note any sensitivities to occlusion or distractors, etc.\par
\vspace{1em}
9. \textbf{Common Failure Modes}\par
List frequently observed failures (e.g., freezing mid-task, grabbing wrong item, failing passive commands). Provide short descriptions and supporting citations.\par
\vspace{1em}
\textbf{Instructions:}\par
\begin{itemize}
    \item When referring to a session, always cite the full session ID (UUID format, e.g., \texttt{16e5bbda-57c1-4e58-a24a-b39ee8142d41}) exactly as provided. Do not shorten, truncate or modify it in any way.
    \item Always wrap session IDs inside \texttt{<ref>...\texttt{</ref>}} tags. Example: \texttt{<ref>16e5bbda-57c1-4e58-a24a-b39ee8142d41</ref>}
    \item Try to cite as many session IDs as possible to support your claims, but only if they are relevant to the point you're making.
    \item Avoid generalizing from a single episode unless there is clear evidence of a pattern.
    \item Keep the tone analytical and professional, emphasizing repeatable behaviors and insights.
    \item Do not invent session IDs. Only use session IDs present in the provided episode reports.
    \item There is no need to mention the specific number of episodes and wins/losses/ties in head-to-head evaluations in this report.
\end{itemize}

\vspace{1em}
The individual episode reports are as follows:\par
\vspace{1em}
\texttt{========== Episode Report \#1 ==========}\par
...
\end{minipage}}
\caption{The prompt template used to generate the full policy performance reports.}
\label{figure:full_policy_performance_analysis_prompt}
\end{figure}

\begin{figure}[ht!]
\colorbox{gray!10}{
\begin{minipage}{\linewidth}
\fontsize{8pt}{9pt}\selectfont
Given the following full evaluation report of a robot manipulation policy, generate a concise, high-quality summary that captures the main findings from sections 1 through 9.\par

\vspace{1em}
Each bullet should summarize the corresponding section in a few sentence fragments, focusing on the most important points. Avoid excessive detail, ensure clarity and correctness. Stick to the facts presented in the full report.\par

\vspace{1em}
Use the following format exactly:\par

\vspace{1em}
- Policy Overview: \textless{}summary\textgreater{}\par

\vspace{1em}
- Comparative Performance: \textless{}summary\textgreater{}\par

\vspace{1em}
- Strengths: \textless{}summary\textgreater{}\par

\vspace{1em}
- Weaknesses: \textless{}summary\textgreater{}\par

\vspace{1em}
- Instruction Following: \textless{}summary\textgreater{}\par

\vspace{1em}
- Reasoning: \textless{}summary\textgreater{}\par

\vspace{1em}
- Manipulation Skills: \textless{}summary\textgreater{}\par

\vspace{1em}
- Robustness to Scene Variations: \textless{}summary\textgreater{}\par

\vspace{1em}
- Common Failure Modes: \textless{}summary\textgreater{}\par

\vspace{1em}
Place a line break between each bullet point. Don't output anything before or after the bullet points.\par

\vspace{1em}
Here is the full report to summarize:\par

\vspace{1em}
\texttt{FULL REPORT}\par
\end{minipage}}
\caption{The prompt template used to generate a concise summary of the full policy performance report.}
\label{figure:summary_policy_performance_analysis_prompt}
\end{figure}

\end{document}